%% file: top.tex
\documentclass[final]{cvpr}

\usepackage[utf8]{inputenc}
\usepackage{times}
\usepackage{multirow}
\usepackage{amsmath,amssymb}
\usepackage{graphicx} 
\usepackage{subcaption}
\usepackage{xspace}
\usepackage{color}
\usepackage{float}
\usepackage{array}
\usepackage[pagebackref=true,breaklinks=true,colorlinks,bookmarks=false]{hyperref}

\def\cvprPaperID{6130}
\def\confYear{CVPR 2021}

\input{shortcuts}

\begin{document}

\title{SMD-Nets: Stereo Mixture Density  Networks}

\author{Fabio Tosi\textsuperscript{1}\thanks{Work done as intern at MPI-IS.} \hspace*{1cm} Yiyi Liao\textsuperscript{2} \hspace*{1cm} Carolin Schmitt\textsuperscript{2} \hspace*{1cm} Andreas Geiger\textsuperscript{2}\\
\textsuperscript{1}Department of Computer Science and Engineering (DISI),
University of Bologna \\ \textsuperscript{2} Autonomous Vision Group, MPI-IS / University of T\"{u}bingen\\
{\tt\small \textsuperscript{1}\{fabio.tosi5\}@unibo.it \textsuperscript{2}\{firstname.lastname\}@tue.mpg.de}}
\clearpage\maketitle
\thispagestyle{empty}
\pagestyle{empty}

\begin{abstract}
Despite stereo matching accuracy has greatly improved by deep learning in the last few years, recovering sharp boundaries and high-resolution outputs efficiently remains challenging. In this paper, we propose Stereo Mixture Density Networks (SMD-Nets), a simple yet effective learning framework compatible with a wide class of 2D and 3D architectures which ameliorates both issues. Specifically, we exploit bimodal mixture densities as output representation and show that this allows for sharp and precise disparity estimates near discontinuities while explicitly modeling the  aleatoric uncertainty inherent in the observations. Moreover, we formulate disparity estimation as a continuous problem in the image domain, allowing our model to query disparities at arbitrary spatial precision. We carry out comprehensive experiments on a new high-resolution and highly realistic synthetic stereo dataset, consisting of stereo pairs at 8Mpx resolution, as well as on real-world stereo datasets. Our experiments demonstrate increased depth accuracy near object boundaries and prediction of ultra high-resolution disparity maps on standard GPUs. We demonstrate the flexibility of our technique by improving the performance of a variety of stereo backbones.
\end{abstract}

\input{1_introduction}
\input{2_related_work}
\input{3_method}
\input{4_experiments}
\input{5_conclusion}

\small{\textbf{Acknowledgements.} This work was supported by the Intel Network on Intelligent Systems, the BMBF through the T\"{u}ebingen AI Center (FKZ: 01IS18039A), the ERC Starting Grant LEGO-3D (850533) and the DFG EXC number 2064/1 - project number 390727645. We thank the IMPRS-IS for supporting Carolin Schmitt. We acknowledge Stefano Mattoccia, Matteo Poggi and Gernot Riegler for their helpful feedback.}

{\small
\bibliographystyle{ieee_fullname}
\bibliography{bibliography_long,bibliography,bibliography_custom}
}

\end{document}

%% file: shortcuts.tex
\newcommand{\bD}{\mathbf{D}}

\newcommand{\bI}{\mathbf{I}}

\newcommand{\bx}{\mathbf{x}}

\newcommand{\nE}{\mathbb{E}}

\newcommand{\nR}{\mathbb{R}}

\newcommand{\cL}{\mathcal{L}}

\DeclareMathOperator*{\argmax}{argmax~}

\makeatletter
\DeclareRobustCommand\onedot{\futurelet\@let@token\@onedot}
\def\@onedot{\ifx\@let@token.\else.\null\fi\xspace}
\def\eg{e.g\onedot} 
\def\ie{i.e\onedot}

\makeatother

\newcommand{\boldparagraph}[1]{\vspace{0.2cm}\noindent{\bf #1:} }

\newcommand{\SEE}{\textit{SEE}\xspace}
\newcommand{\EPE}{\textit{EPE}\xspace}
\newcommand{\SEEk}[1]{\textit{SEE}_{#1}}

\definecolor{darkcyan}{rgb}{0.0, 0.75, 0.75}
\definecolor{darkgreen}{rgb}{0,0.7,0}
\definecolor{orange}{rgb}{1,0.5,0}



%% file: 1_introduction.tex
\section{Introduction}

Stereo matching is a long standing and active research topic in computer vision. It aims at recovering dense correspondences between  image pairs by estimating the \textit{disparity} between matching pixels, required to infer depth through triangulation. It also plays a crucial role in many areas like 3D mapping, scene understanding and robotics. 

Traditional stereo matching algorithms apply hand-crafted matching costs and engineered regularization strategies. More recently, learning methods based on Convolutional Neural Networks (CNNs) have proven to be superior, given the increasing availability of large stereo datasets \cite{geiger2012we, geiger2013vision, yang2019drivingstereo}. Although such methods produce compelling results, two major issues remain unsolved: predicting accurate depth boundaries and generating high-resolution outputs with limited memory and computation.

\begin{figure}[t]
  \centering
  \renewcommand{\tabcolsep}{1pt} 
 	\begin{subfigure}{\linewidth}
        \includegraphics[width=0.95\linewidth]{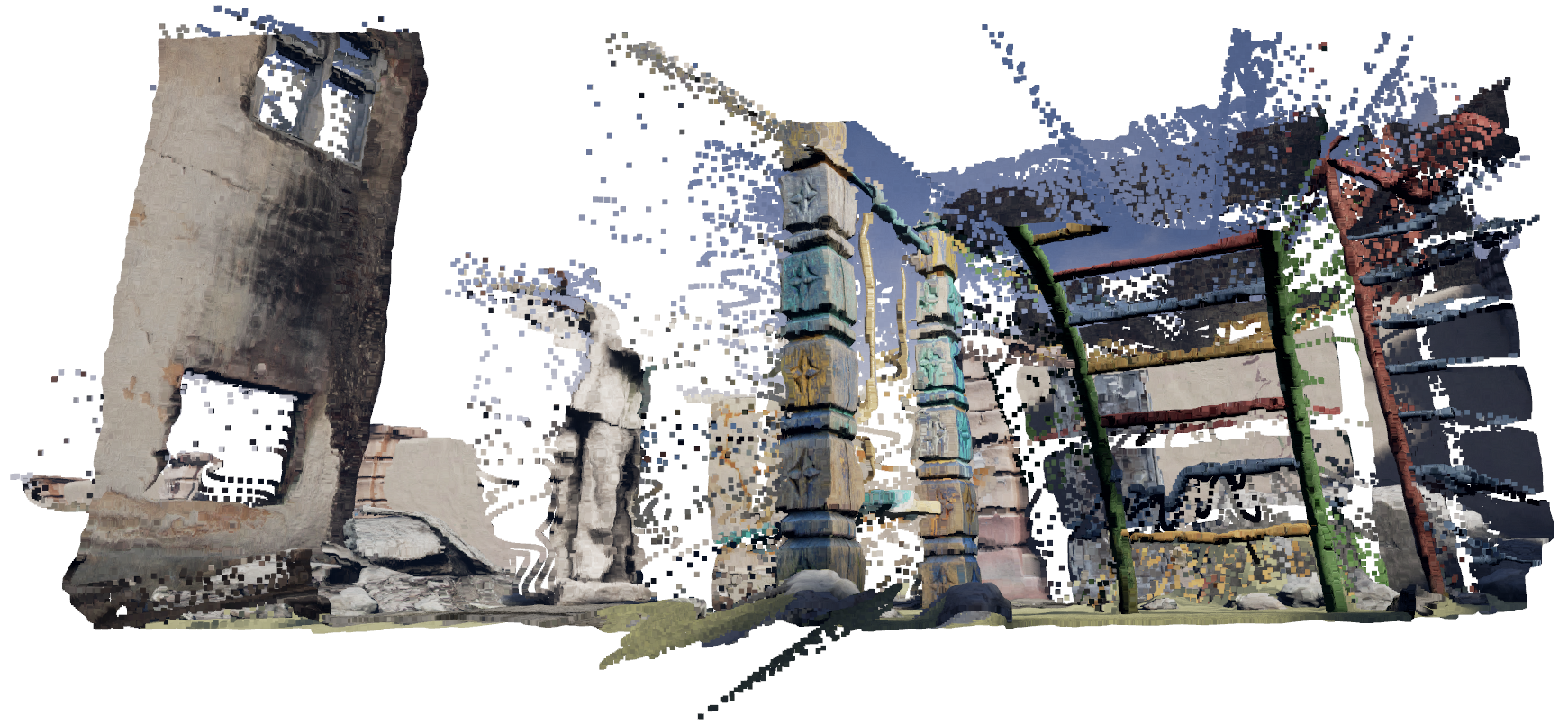}
         \caption{ PSM \cite{chang2018pyramid}}
         \label{fig:teaser_psm}
    \end{subfigure}
    \begin{subfigure}{\linewidth}
        \includegraphics[width=0.95\linewidth]{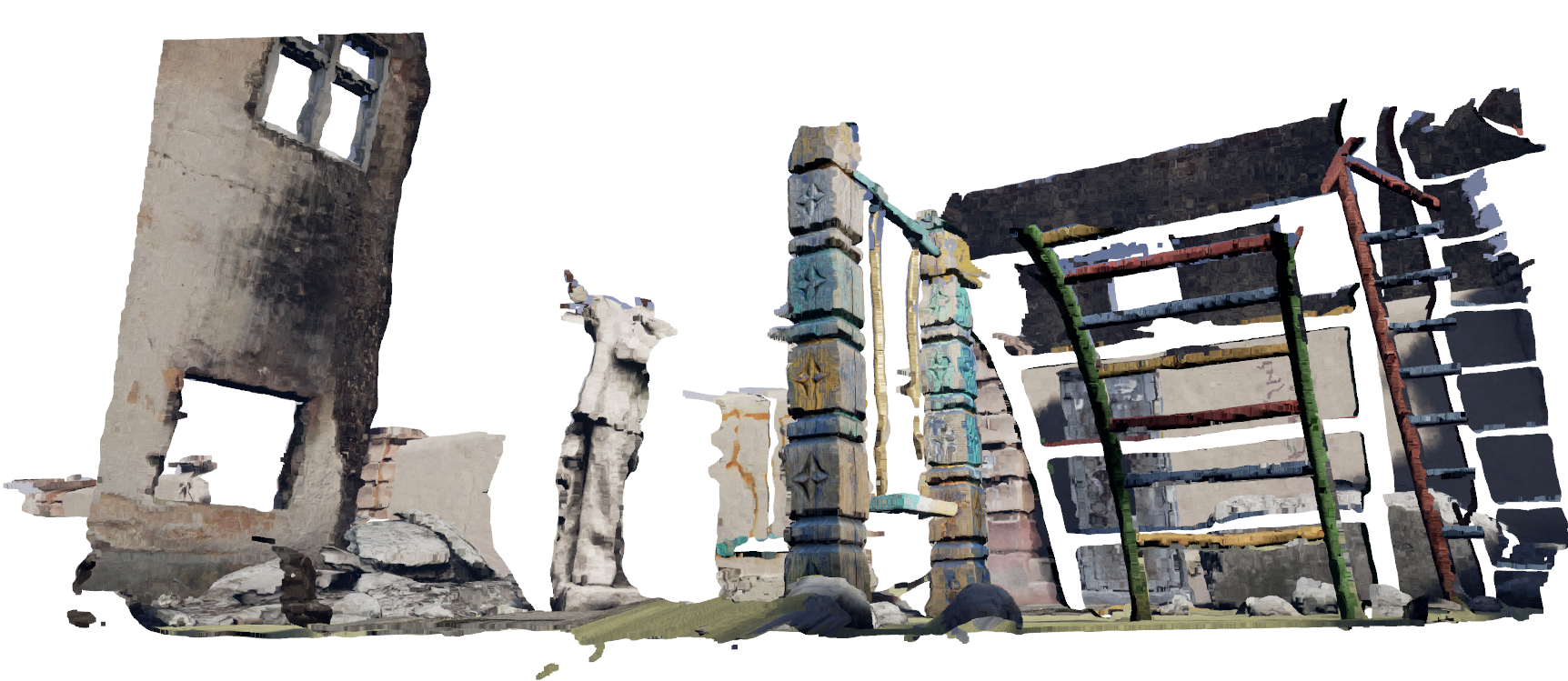}
         \caption{PSM \cite{chang2018pyramid} + Ours}
         \label{fig:teaser_ours}

    \end{subfigure}
    \caption{\textbf{Point Cloud Comparison} between the stereo network PSM \cite{chang2018pyramid} and our Stereo Mixture Density Network (SMD-Net) on the UnrealStereo4K dataset. Notice how SMD-Net notably alleviates bleeding artifacts near object boundaries, resulting in more accurate 3D reconstructions. }

  \label{fig:teaser}
\end{figure}

The first issue is shown in  \figref{fig:teaser_psm}: As neural networks are smooth function approximators, they often poorly reconstruct object boundaries, causing ``bleeding'' artifacts (i.e., flying pixels) when converted to point clouds. These artifacts can be detrimental to subsequent applications such as 3D reconstruction or 3D object detection.  Thus, while being ignored by most commonly employed disparity metrics, accurate 3D reconstruction of contours is a desirable property for any stereo matching algorithm.

Furthermore, existing methods are limited to discrete predictions at pixel locations of a fixed resolution image grid, while geometry is a piecewise continuous quantity where object boundaries may not align with pixel centers. Increasing the output resolution by adding extra upsampling layers partially addresses this problem as this leads to a significant increase in memory and computation.

In this work, we address both issues. Our key contribution is to learn a representation that is precise at object boundaries and scales to high output resolutions. In particular, we formulate the task as a continuous estimation problem and exploit bimodal mixture densities \cite{Bishop1994} as output representation. Our simple formulation lets us (1) avoid bleeding artifacts at depth discontinuities, (2) regress disparity values at arbitrary spatial resolution with constant memory and (3) provides a measure for aleatoric uncertainty. 

\begin{figure*}[h]
    \centering
	\begin{subfigure}{0.45\linewidth}
		\includegraphics[width=\linewidth]{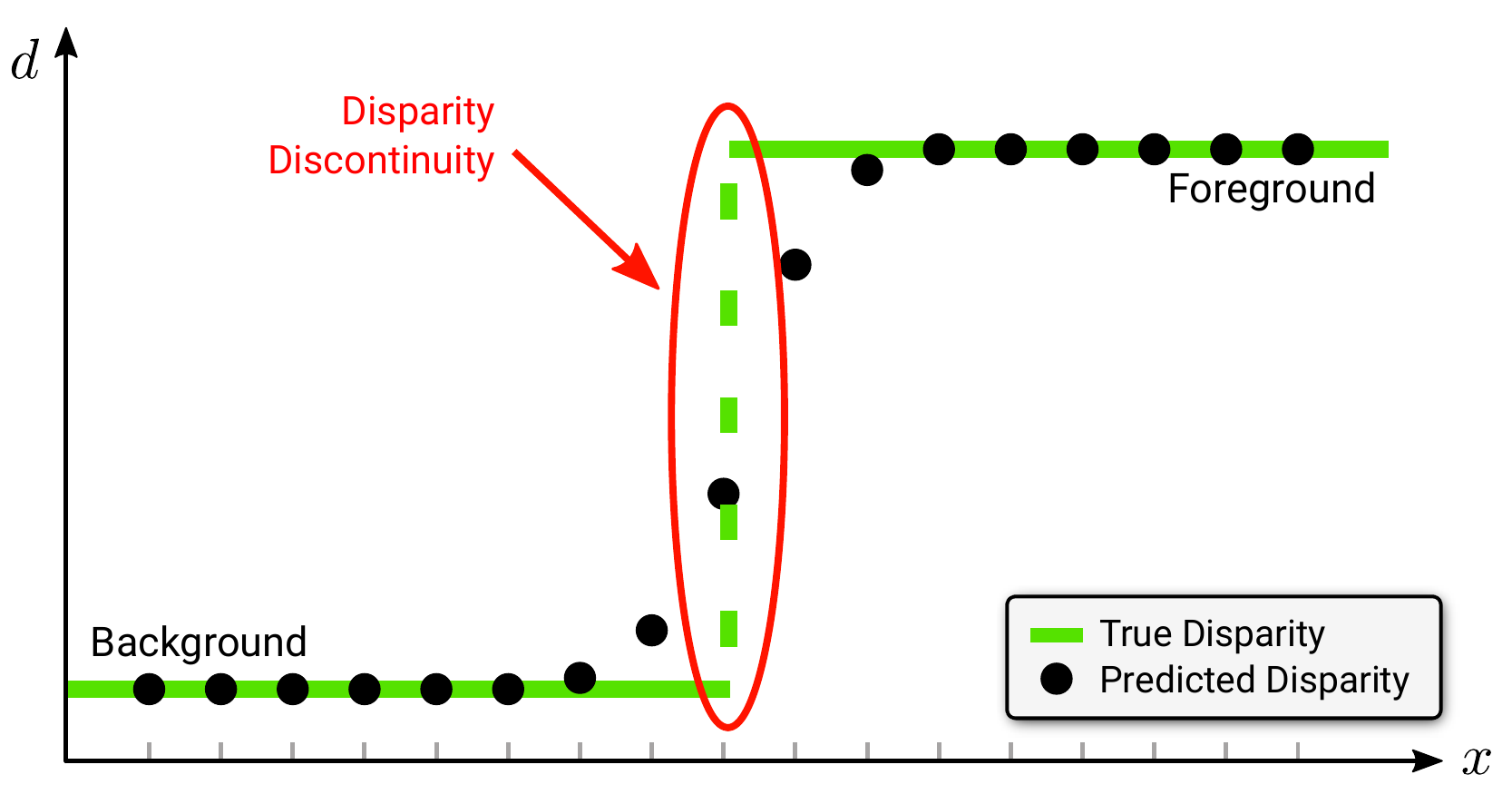}
		\caption{Stereo Regression Network}
		\label{fig:illustration_stereo_net}
	\end{subfigure}
	\quad
	\begin{subfigure}{0.45\linewidth}
		\includegraphics[width=\linewidth]{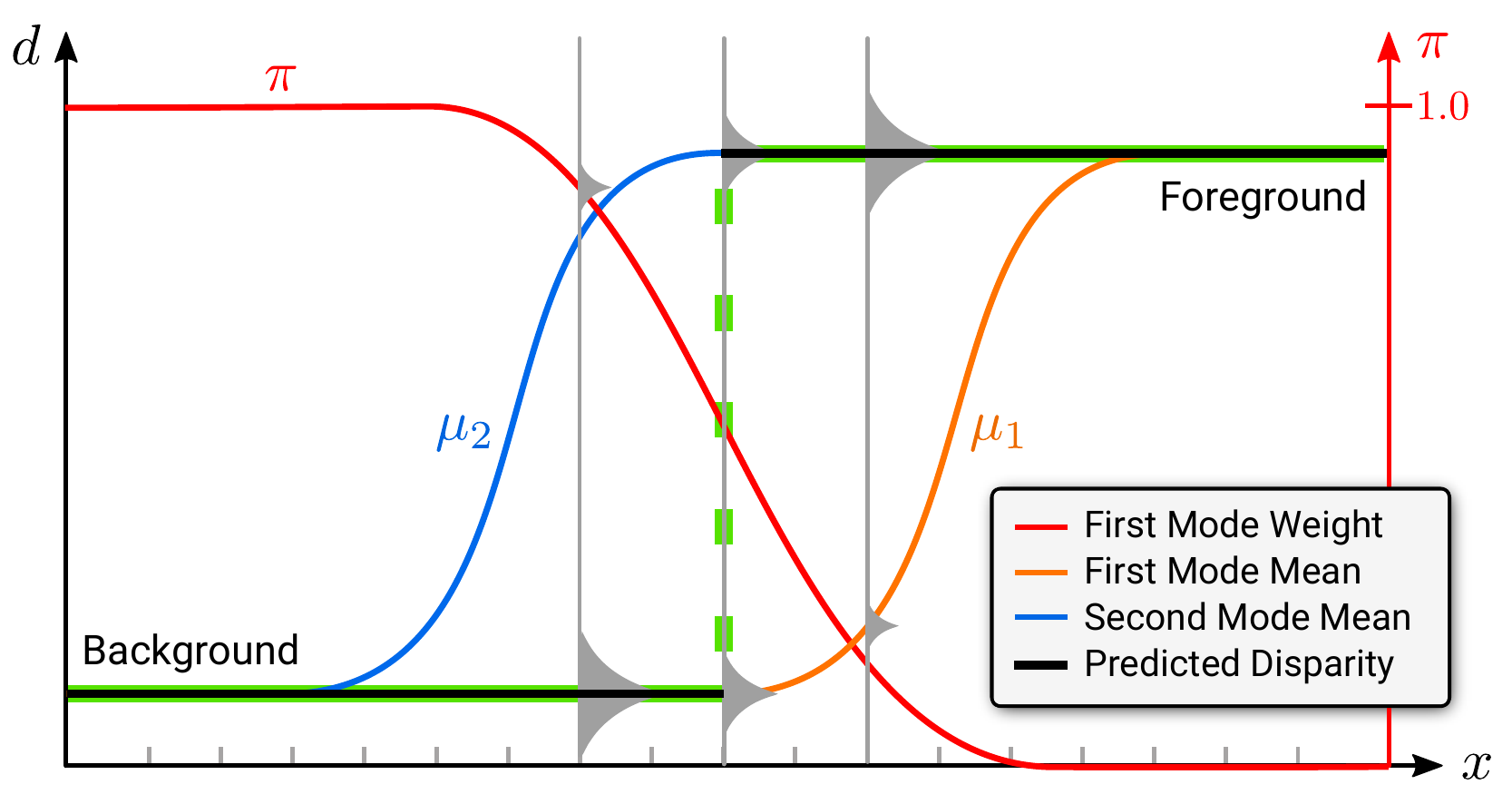}
		\caption{Stereo Mixture Density  Network}
		\label{fig:illustration_mds_net}
	\end{subfigure}
	\vspace{-0.2cm}
	\caption{{\bf Overcoming the Smoothness Bias with Mixture Density Networks.} For clarity, we visualize the disparity $d$ only for a single image row. (\subref{fig:illustration_stereo_net}) Classical deep networks for stereo regression suffer from smoothness bias and hence continuously interpolate object boundaries. In addition, disparity values are typically predicted at discrete spatial locations. (\subref{fig:illustration_mds_net}) In this work, we propose to use a bimodal Laplacian mixture distribution (illustrated in gray) with weight $\pi$ as output representation which can be queried at any continuous spatial location $x$. This allows our model to accurately capture uncertainty close to depth discontinuities while at inference time recovering sharp edges by selecting the mode with the highest probability density.  In this example,  the first mode ($\mu_{1}, b_{1}$) models the background and the second mode ($\mu_{2}, b_{2}$) models the foreground disparity close to the discontinuity. When the probability density of the foreground mode becomes larger than the probability density of the background mode, the most likely disparity sharply transitions from the background to the foreground value. 
	\vspace{-0.2cm}
	}
	\label{fig:illustration}
\end{figure*}

We illustrate the boundary bleeding problem and our solution to it in \figref{fig:illustration}. While classical deep networks for stereo regression suffer from smoothness bias and are incapable of representing sharp disparity discontinuities, the proposed Stereo Mixture Density Networks (SMD-Nets) effectively address this issue. The key idea is to alter the output representation adopting a mixture distribution such that sharp discontinuities can be regressed \textit{despite} the fact that the underlying neural networks are only able to make smooth predictions (note that all curves in \figref{fig:illustration_mds_net} are indeed smooth while the predicted disparity is discontinuous). 

Furthermore, the proposed model is capable of regressing disparity values at arbitrary continuous locations in the image,  effectively solving a stereo super-resolution task.
In combination with the proposed representation, this allows for regressing sharp discontinuities at sub-pixel resolution while keeping memory requirements constant.

In summary, we present: (i) A novel learning framework  for stereo matching that exploits compactly parameterized bimodal mixture densities as output representation and can be trained using a simple likelihood-based loss function. (ii) A continuous function formulation aimed at estimating disparities at arbitrary spatial resolution with constant memory footprint. (iii) A new large-scale synthetic binocular stereo dataset with ground truth disparities at $3840 \times 2160$ resolution, comprising photo-realistic renderings of indoor and outdoor environments. (iv) Extensive experiments on several datasets demonstrating improved accuracy at depth discontinuities for various backbones on binocular stereo, monocular and active depth estimation tasks.

Our source code and dataset are available at \url{https://github.com/fabiotosi92/SMD-Nets}.

%% file: 2_related_work.tex
\section{Related Work}

\vspace{-0.2cm}
\boldparagraph{Deep Stereo Matching} 
Stereo has a long history in computer vision \cite{scharstein2002taxonomy}. 
With the rise of deep learning, CNN based methods for stereo were pioneered in \cite{vzbontar2016stereo} with the aim of replacing the traditional matching cost computation. 

More recent works attempt to solve the stereo matching task without hand-crafted post processing steps. They can be categorized into 2D architectures and 3D architectures.  In the first category, \cite{liang2018learning, pang2017cascade, ilg2018occlusions, yang2018segstereo, yin2019hierarchical, tonioni2019real, aleotti2019learning} extend the seminal DispNet~\cite{mayer2016large}, an end-to-end network for disparity regression.
The second class, instead, consists of architectures that explicitly construct 3D feature cost volumes by means of concatenation/feature difference \cite{chang2018pyramid, khamis2018stereonet, tulyakov2018practical, wang2019anytime, duggal2019deeppruner, zhang2019ga, wu2019semantic, you2019pseudo, nie2019multi, chabra2019stereodrnet, dovesi2020real, song2020edgestereo, xu2020aanet, kusupati2020normal} and group-wise correlation \cite{guo2019group}. 
A thorough review of these works can be found in \cite{poggi2020synergies}. We stress once again how such networks, although achieving state-of-the-art results on most stereo benchmarks, suffer from severe over-smoothing at object discontinuities which are not captured by commonly employed disparity metrics, but which matter for many downstream applications.
Therefore, the ideas proposed in this work to address this issue are orthogonal to the aforementioned networks and can be advantageously combined with nearly any stereo backbone.

\boldparagraph{Disparity Output Representation}
Standard stereo networks directly regress a scalar disparity at every pixel. This output representation suffers from over-smoothing and does not expose the underlying aleatoric uncertainty.
The latter problem can be addressed by modeling the disparity using a \textit{parametric} distribution,
e.g., a Gaussian or Laplacian distribution \cite{kendall2017uncertainties,mehltretter2019cnn} while the over-smoothing issue remains unsolved. 
A key result of our work is to demonstrate that replacing the unimodal output representation with a bimodal one is sufficient to significantly alleviate this problem.

Another line of methods estimate a \textit{non-parametric} distribution over a set of discrete disparity values.
However, this approach leads to inaccurate results when the estimated distribution is multi-modal \cite{kendall2017end}. Some works tackle the problem by enforcing a unimodal constraint during training \cite{chen2019over, zhang2019adaptive}. In contrast, we explicitly model the bimodal nature of the distribution at object boundaries by adopting a simple and effective bimodal representation. In concurrent work, \cite{Garg2020WassersteinDF} also predicts multi-modal distributions supervised by a heuristically designed multi-modal ground truth over a set of depth values. In contrast to them, our bimodal approach can be learned by maximizing the likelihood without requiring direct supervision on the distribution itself.

\boldparagraph{Continuous Function Representation}
Existing deep stereo networks use fully convolutional neural networks and make predictions at discrete pixel locations. 
Recently, continuous function representations have gained attention in many areas, including 3D reconstruction~\cite{Mescheder2019CVPR,Park2019CVPR,Chen2019CVPR,Saito2019ICCV,Sitzmann2020NIPS,Niemeyer2020CVPR,Peng2020ECCV}, texture estimation~\cite{Oechsle2019ICCV}, image synthesis~\cite{Mildenhall2020ECCV,Schwarz2020NIPS} and semantic segmentation~\cite{Kirillov2020CVPR}. 
To the best of our knowledge, we are the first to adopt a continuous function representation for disparity estimation, allowing us to predict a disparity value at any continuous pixel location. In contrast to works that allow for high-resolution stereo matching by designing memory efficient architectures~\cite{yang2019hierarchical, gu2020cascade}, our simple output representation is able to exploit ground truth disparity maps at a higher resolution than the input stereo pair, thus effectively learning stereo super-resolution. 

%% file: 3_method.tex
\section{Method}

\begin{figure*}[t]
    \centering
	\includegraphics[width=\linewidth]{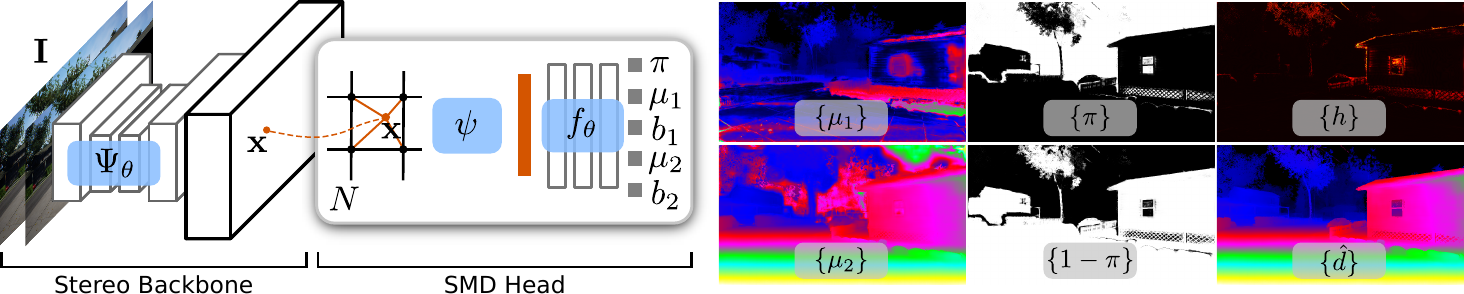}
	\caption{{\bf Method Overview.} We assume a 2D or 3D stereo backbone network $\Psi_{\theta}$ which takes as input a stereo pair $\bI$ (either concatenated or processed by siamese towers), and outputs a $D$-dimensional feature map in the domain of the reference image. 
	Given any continuous 2D location $\bx$, we query its feature from the feature map via bilinear interpolation as denoted by $\psi$. The interpolated feature vector is then fed into a multi-layer perceptron $f_{\theta}$ to estimate a five-dimensional vector ($\pi,\mu_{1}, b_{1},\mu_{2}, b_{2}$) which represents the parameters of a bimodal distribution.
	$N$ denotes the number of points randomly sampled at continuous 2D locations during training and the number of pixels during inference. On the right we show maps of $\mu_{1}$, $\mu_{2}$, $\pi$, $1-\pi$, uncertainty $h$ and predicted disparity $\hat{d}$.
	}
	\vspace{-0.2cm}
	\label{fig:architecture}
\end{figure*}

\figref{fig:architecture} illustrates our model. We first encode a stereo pair into a feature map using a convolutional backbone (left). Next, we estimate parameters of a mixture density distribution at any continuous 2D location via a multi-layer perceptron head, taking the bilinearly interpolated feature vector as input (middle). From this, we obtain a disparity as well as uncertainty map (right). We now explain our model, loss function and training protocol in detail.

\subsection{Problem Statement}
Let $\bI\in\nR^{W \times H \times 6}$ denote an RGB stereo pair for which we aim to predict a disparity map $\bD$ at arbitrary resolution. As shown in \figref{fig:illustration}, classical stereo regression networks suffer from over-smoothing due to the smoothness bias of neural networks. In this work, we exploit a mixture distribution as output representation \cite{Bishop1994} to overcome this limitation. 

More specifically, we propose to use a bimodal Laplacian mixture distribution with weight $\pi$ and two modes $(\mu_{1}, b_{1})$, $(\mu_{2}, b_{2})$ to model the continuous probability distribution over disparities at a particular pixel. 
Using two modes allows our model to capture both the foreground as well as the background disparity at object boundaries. 
At inference time, we recover sharp object boundaries by selecting the mode with the highest density value. Thus, our model is able to transition from one disparity to another in a discontinuous fashion while at the same time relying only on the regression of functions ($\pi,\mu_{1}, b_{1},\mu_{2}, b_{2}$) which are smooth with respect to the image domain and which therefore can easily be represented using neural networks.

\subsection{Stereo Mixture Density Networks} We now formally describe our model. Let
\begin{equation}
\Psi_\theta: \nR^{W \times H \times 6} \rightarrow \nR^{W \times H \times D}
\end{equation}
denote a \textit{stereo backbone} network with parameters $\theta$ as shown in \figref{fig:architecture} (left). $\Psi_\theta$ takes as input the stereo pair $\bI$ and outputs a $D$-dimensional feature map, represented in the domain of the reference image (\eg the left image of a stereo pair). Examples for such networks are standard 2D convolutional networks, or networks which perform 3D convolutions. For the 2D networks, the stereo pair can be concatenated as input or processed by means of siamese towers with shared weights as typically done for 3D architectures. Similarly, this generic formulation also applies to the structured light setting (\eg, Kinect setting where $\bI \in \nR^{W\times H}$) and the monocular depth estimation problem ($\bI \in \nR^{W \times H \times 3}$).

As geometry is a piecewise continuous quantity, we apply a deterministic transformation to obtain feature points for any continuous location in $\nR^{W \times H}$. More specifically, for every continuous 2D location $\bx \in \nR^2$, we bilinearly interpolate the features from its four nearest pixel locations in the feature map $\nR^{W \times H \times D}$. More formally, we describe this transformation as: 
\begin{equation}
\psi: \nR^2 \times \nR^{W \times H \times D} \rightarrow \nR^D
\end{equation}
Finally, we employ a multi-layer perceptron to map this abstract feature representation to a five-dimensional vector ($\pi,\mu_{1}, b_{1},\mu_{2}, b_{2}$) which represents the parameters of a univariate bimodal mixture distribution:
\begin{equation}
f_\theta: \nR^D \rightarrow \nR^5
\end{equation}
Note that we have re-used the parameter symbol $\theta$ to simplify notation. In the following, we use $\theta$ to denote all parameters of our model. We refer to $f_\theta(\psi(\cdot,\cdot))$ as \textit{SMD Head}, see \figref{fig:architecture} for an illustration.

\begin{figure}
    \centering
    \includegraphics[width=\linewidth]{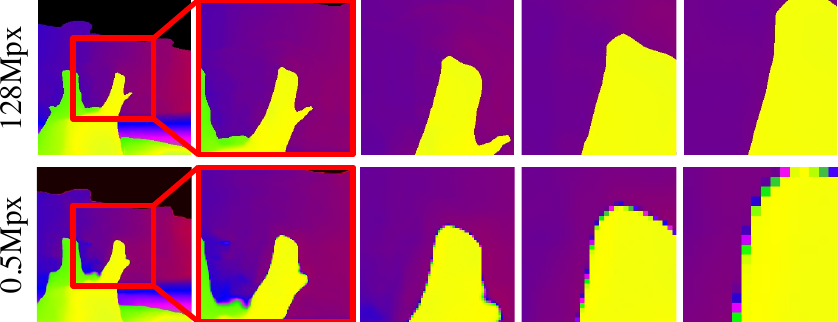}
    \vspace{-0.3cm}
    \caption{\textbf{Ultra High-resolution Estimation.} 
    Comparison of our model using the PSM backbone at 128Mpx resolution (top) to the original PSM at 0.5Mpx resolution (bottom), both taking stereo pairs at 0.5Mpx resolution as input. Each column shows a different zoom-level. Note how our method leads to sharper boundaries and high resolution outputs.}
    \label{fig:ultra-res}
    \vspace{-0.2cm}
\end{figure}

To robustly model a distribution over disparities which can express two modes close to disparity discontinuities, we choose a bimodal Laplacian mixture as output representation:
\begin{equation} 
p(d)=
\frac{\pi}{2\,b_{1}}e^{-\frac{|d-\mu_{1}|}{b_{1}}} + \frac{1-\pi}{2\,b_{2}}e^{-\frac{|d-\mu_{2}|}{b_{2}}}
\label{eq:laplacian_mixture}
\end{equation}
In summary, our model can be compactly expressed as:
\begin{equation}
p(d|\bx,\bI,\theta) = p(d|f_\theta \left(\psi(\bx,\Psi_\theta(\bI)))\right)
\end{equation}
At inference time, we determine the final disparity $\hat{d}$ by choosing the mode with the highest density value:
\begin{equation}
\hat{d} = \argmax_{d\in \{\mu_{1}, \mu_{2}\}} p(d)
\end{equation}
Note that our formulation allows to query the disparity $\hat{d} \in \mathbb{R}$ at any continuous 2D pixel location, enabling ultra high-resolution predictions  with sharply delineated object boundaries. This is illustrated in \figref{fig:ultra-res}.

Our model also allows for capturing the aleatoric uncertainty of the predicted disparity by evaluating the differential entropy of the continuous mixture distribution as:
\begin{equation}
h = -\int p(d) \log p(d) \,\mathrm{d}d
\end{equation}
In practice, we use numerical quadrature to obtain an approximation of the integral. 

\subsection{Loss Function}

We consider the supervised setting and train our model by minimizing the negative log-likelihood loss:

\begin{equation}
\cL_{NLL}(\theta) = - \,\nE_{d,\bx,\bI} \, \log p(d|\bx,\bI,\theta)
\end{equation}
where the input $\bI$ is randomly sampled from the dataset, $\bx$ is a random pixel location in the continuous image domain $\Omega = [0,W-1] \times [0,H-1]$, sampled as described in \secref{sec:training_protocol}, and $d$ is the ground truth disparity at location $\bx$.

\subsection{Training Protocol}\label{sec:training_protocol}

\boldparagraph{Sampling Strategy}
While a na\"{i}ve strategy samples pixel locations $\bx$ randomly and uniformly from the image domain $\Omega$, our framework also allows for exploiting custom sampling strategies to focus on depth discontinuities during training.
We adopt a \textit{Depth Discontinuity Aware (DDA)} sampling approach during training that explicitly favors points located near object boundaries while at the same time maintaining a uniform coverage on the entire image space. 
More specifically, given a ground truth disparity map at training time, we first compute an object boundary mask in which a pixel is considered to be part of the boundary if its (4-connected) neighbors have a disparity that differs by more than 1 from its own disparity. This  mask is then dilated using a $\rho \times \rho$ kernel to enlarge the boundary region. We report an analysis using different $\rho$ values in the experimental section.
Given the total number of training points $N$, we randomly and uniformly select $N/2$ points from the domain of all pixels belonging to depth discontinuity regions and $N/2$ points uniformly from the continuous domain of all remaining pixels.
At inference time,  we leverage our model to predict disparity values at each location of an (arbitrary resolution) grid. 

\boldparagraph{Stereo Super-Resolution} Our continuous formulation allows us to exploit ground truth at higher resolution than the input $\bI$, which we refer to as stereo super-resolution. In contrast, classical stereo methods cannot realize arbitrary super-resolution without changing their architecture.

%% file: 4_experiments.tex
\section{Experimental Results}
\label{experiments}

In this section, we first describe the datasets used for evaluation and implementation details. We then present an extensive evaluation that demonstrates the benefits of the proposed SMD Head in combination with different stereo backbones on several distinct tasks.

\subsection{Datasets}

\boldparagraph{UnrealStereo4K} Motivated by the lack of large-scale, realistic and high-resolution stereo datasets, we introduce a new photo-realistic \textit{binocular stereo} dataset at $3840\times 2160$ resolution with pixel-accurate ground truth. We create this synthetic dataset using the popular game engine Unreal Engine combined with the open-source plugin UnrealCV~\cite{Qiu2017ACM}. We additionally create a synthetic \textit{active monocular} dataset (mimicking the Kinect setup) at $4112\times 3008$ resolution by warping a gray-scale reference dot pattern to each image, following \cite{Riegler2019CVPR}. We split the dataset into $7720$ training pairs, $80$ validation pairs and $200$ \textit{in-domain} test pairs. To evaluate the generalization ability of our method, we also create an \textit{out-of-domain} test set by rendering $200$ stereo pairs from an unseen scene.
Similarly, the active dataset contains $3856$ training images, $40$ validation images, $100$ test images.

\boldparagraph{RealActive4K} We further collect a small real-world active dataset of an indoor room with a Kinect-like stereo sensor, including 2570 images at a resolution $4112 \times 3008$ pixels from which we use $2500$ for training, $20$ for validation and $50$ for testing. 
We perform Block Matching with left-right consistency check to use as co-supervision for training models jointly on synthetic (UnrealStereo4K) and real data.

\boldparagraph{KITTI 2015 \cite{menze2015object}} The KITTI dataset is a collection of real-world stereo images depicting driving scenarios. It contains $200$ training pairs with sparse ground truth depth maps collected by a LiDAR and $200$ testing pairs. We divide the KITTI training set into $160$ training stereo pairs and $40$ validation stereo pairs, following \cite{song2020edgestereo}.
\medskip

\boldparagraph{Middlebury v3 \cite{scharstein2014high}} Middlebury v3 is a small high-resolution stereo dataset depicting indoor scenes under controlled lighting conditions containing $10$ training pairs and $10$ testing pairs with dense ground truth disparities.


\subsection{Implementation Details }
\boldparagraph{Architecture} In principle, our SMD Head is compatible with any stereo backbone $\psi_\theta$ from the literature. In our implementation, we build on top of two state-of-the-art 3D stereo architectures: Pyramid Stereo Matching (PSM) network \cite{chang2018pyramid} and Hierarchical Stereo Matching (HSM) network \cite{yang2019hierarchical}. PSM is a well-known and popular stereo network while HSM represents a method with good trade-off between accuracy and computation. Moreover, we also adopt a na\"ive U-Net structure \cite{monodepth2} that takes as input concatenated images of a stereo pair in order to show the effectiveness of our model on 2D architectures. For the aforementioned networks, we follow the official code provided by the authors.

Our SMD Head $f_{\theta}$ is implemented as a multi-layer perceptron (MLP) following \cite{Saito2019ICCV}. More specifically, the number of neurons is $(D, 1024, 512, 256, 128, 5)$. We use sine activations \cite{Sitzmann2020NIPS} except for the last layer that uses a sigmoid activation for regressing  the  five-parameter  output. For the 3D backbone, we select the matching probabilities from the cost volume in combination with features of $\Psi_{\theta}$ at different resolutions as input to our SMD Head. For the 2D backbone case, instead, we select features from different layers of the decoder. We refer the reader to the supplementary material for details.

\boldparagraph{Training} We implement our approach in PyTorch \cite{paszke2019pytorch} and use Adam  with $\beta_1= 0.9$  and $\beta_2= 0.999$ as optimizer \cite{Kingma2015ICLR}. We train all models from scratch using a single NVIDIA V100 GPU. During training, we use random crops from $\bI$ as input to the stereo backbone and sample $N=50,000$ training points from each crop. We scale the ground truth disparity to $[0,1]$ for each dataset for numerical stability.
Moreover, for RGB inputs we perform chromatic augmentations on the fly, including random brightness, gamma and color shifts sampled from uniform distributions. We further apply horizontal and vertical random flipping while adapting the ground truth disparities accordingly. 
Please see the supplementary material for details regarding the training procedure for each dataset.

\boldparagraph{Evaluation Metrics} Following \cite{chen2019over}, we evaluate the \textit{Soft Edge Error ($\textit{SEE}_k$)} metric on pixels belonging to object boundaries, defined as the minimum absolute error between the predicted disparity and the corresponding local ground truth patch of size $k \times k$ ($k=\{3,5\}$ in our experiments). Intuitively, \SEE penalizes over-smoothing artifacts stronger compared to small misalignments in a local window, where the former is more harmful to subsequent applications.

While not our main focus, we also report the \textit{End Point Error (EPE)} as the standard error metric obtained by averaging the absolute difference between predictions and ground truth disparity values to evaluate the overall performance. For both \SEE and \EPE, we compute the average (Avg) and $\sigma(\Delta)$ metrics, with the latter one representing the percentage of pixels having errors greater  than $\Delta$.

\subsection{Ablation Study}

We first examine the impact of different components and  training choices of the proposed SMD-Nets on the \textit{in-domain} UnrealStereo4K test set.
Unless specified otherwise, we use $960\times540$ as resolution for the binocular input $\bI$ and $3840\times2160$ for the corresponding ground truth, used for both supervision and testing purposes. The active input images consist of random dot patterns where the dots become indistinguishable at low resolution (\eg, $960\times540$). Therefore we use $2056\times 1504$ as active input size while keeping the ground truth dimension at $4112\times 3008$.

\boldparagraph{Output Representation} In \tabref{table:ablation}, we evaluate the effectiveness of our mixture density output representation across both, 2D and 3D stereo backbones on multiple tasks including binocular stereo, monocular depth and active depth.
We adopt U-Net and PSM on the binocular stereo dataset as representatives of 2D and 3D backbones and report results of HSM in the supplementary for the sake of space. We also use the same U-Net backbone for a monocular depth estimation task by replacing the input with only the reference image of a binocular stereo pair to show the advantage of our method on various tasks. For the active setup, we choose HSM as it represents a network designed specifically for high-resolution inputs which takes as input the monocular active image and the \textit{fixed} reference dot pattern. 

We compare our bimodal distribution to two other output representations, standard disparity regression and a unimodal Laplacian distribution \cite{kendall2017uncertainties}.
For fairness, we implement these baselines by replacing the last layer of our SMD Head to predict the disparity $d$ or the unimodal parameters $(\mu, b)$, respectively, where the former is trained with a standard L1 loss while the latter with a negative log-likelihood loss. For all cases we use the proposed bilinear feature interpolation and the na\"ive random sampling strategy.

\tabref{table:ablation} shows that the proposed method effectively addresses the over-smoothing problem at object boundaries, achieving the lowest \SEE for all backbones on all tasks, compared to both the standard disparity regression and the unimodal representation. Moreover, we observe that the unimodal representation sacrifices \EPE for capturing the uncertainty, while our method is on par with the standard L1 regression.
On the stereo dataset, the 3D backbone (PSM) consistently outperforms the 2D backbone (U-Net), therefore we use PSM for the following ablation experiments.

\input{tables/unreal_ablation2}

\input{tables/dilation_factor}

\input{tables/superresolution}

\input{tables/unreal_comparison}

\begin{figure*}
    \centering
    \renewcommand{\tabcolsep}{1pt}   
    \begin{tabular}{cccc}
        \includegraphics[width=0.24\textwidth]{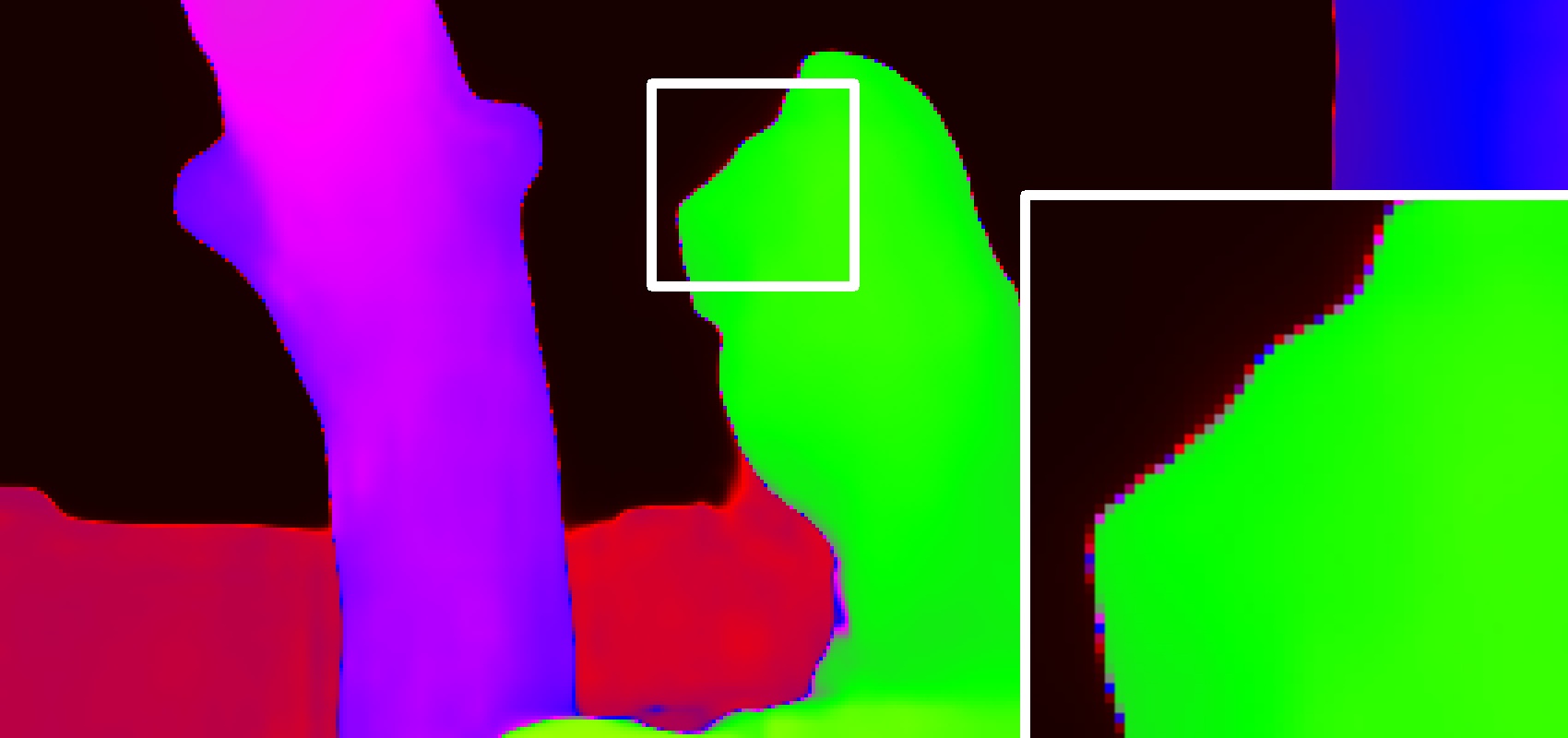}   &
        \includegraphics[width=0.24\textwidth]{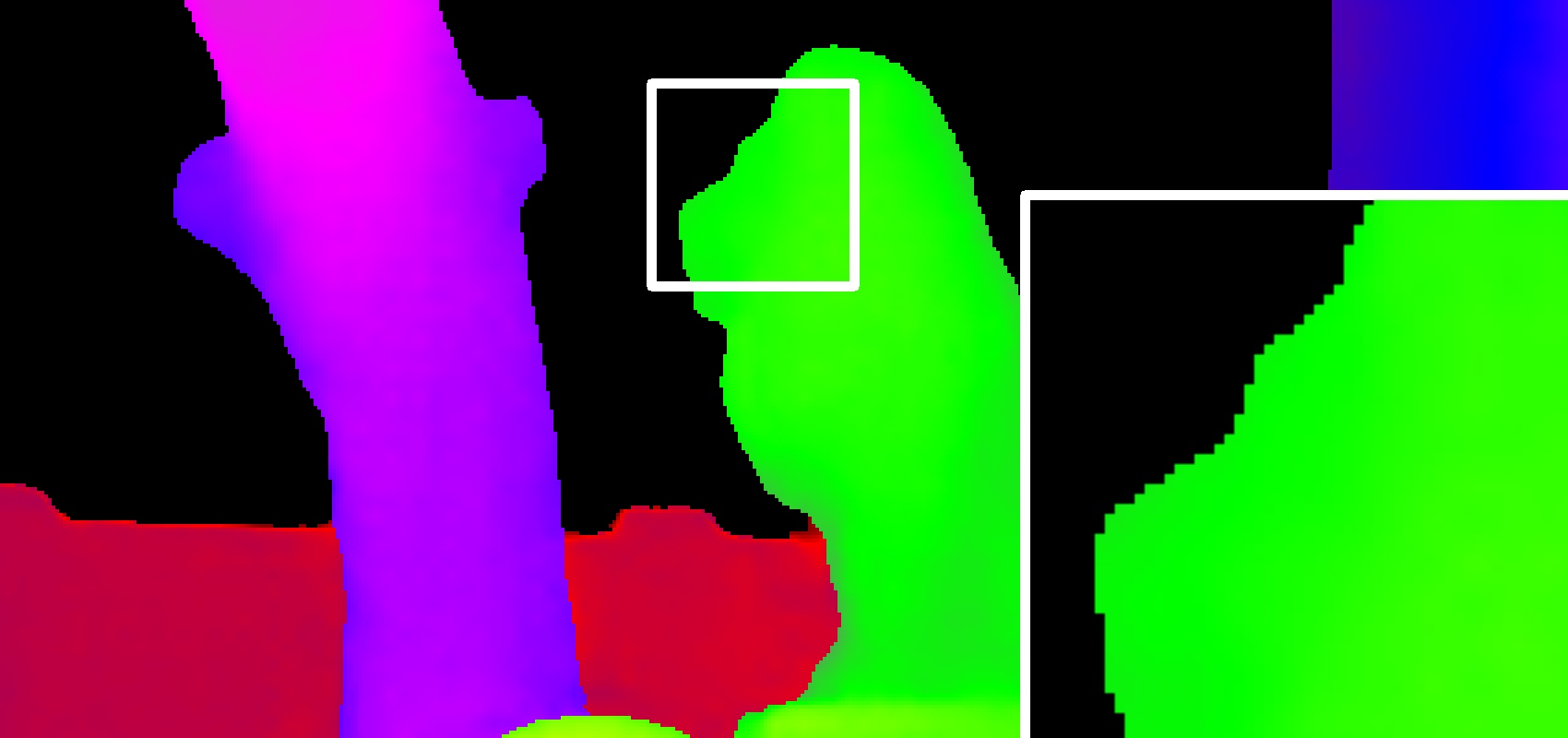}   &
        \includegraphics[width=0.24\textwidth]{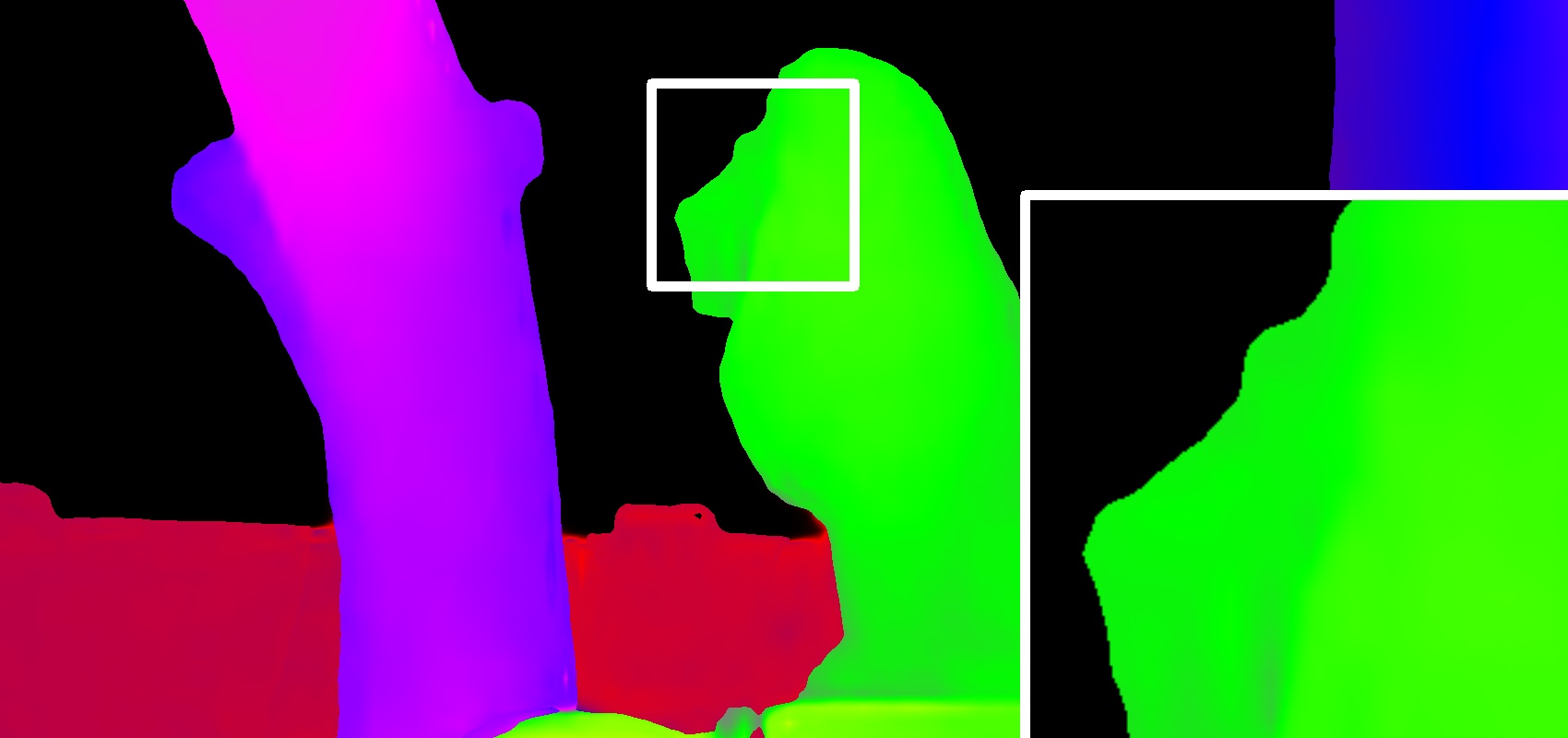} &
        \includegraphics[width=0.24\textwidth]{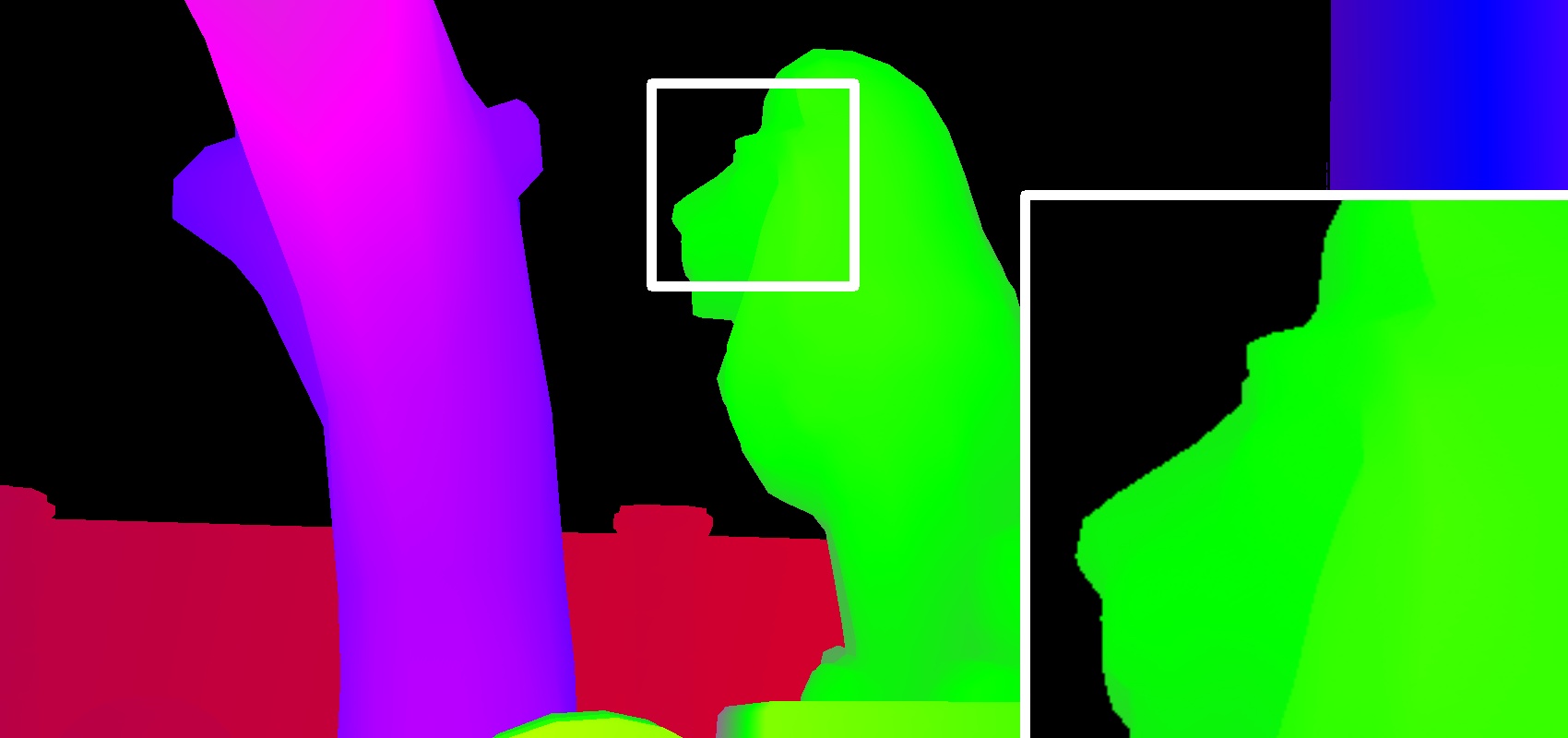} 
        \\
        \includegraphics[width=0.24\textwidth]{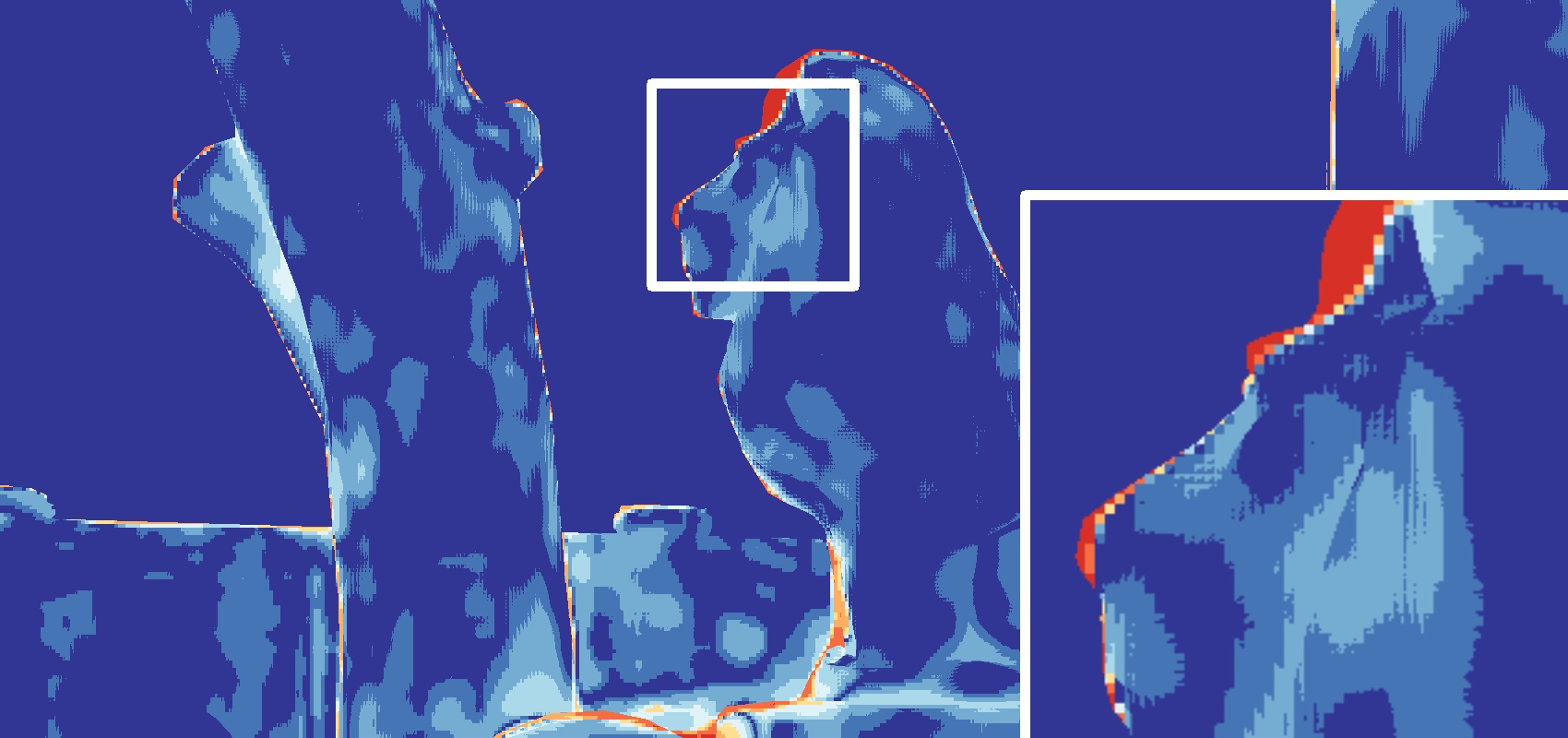} &
        \includegraphics[width=0.24\textwidth]{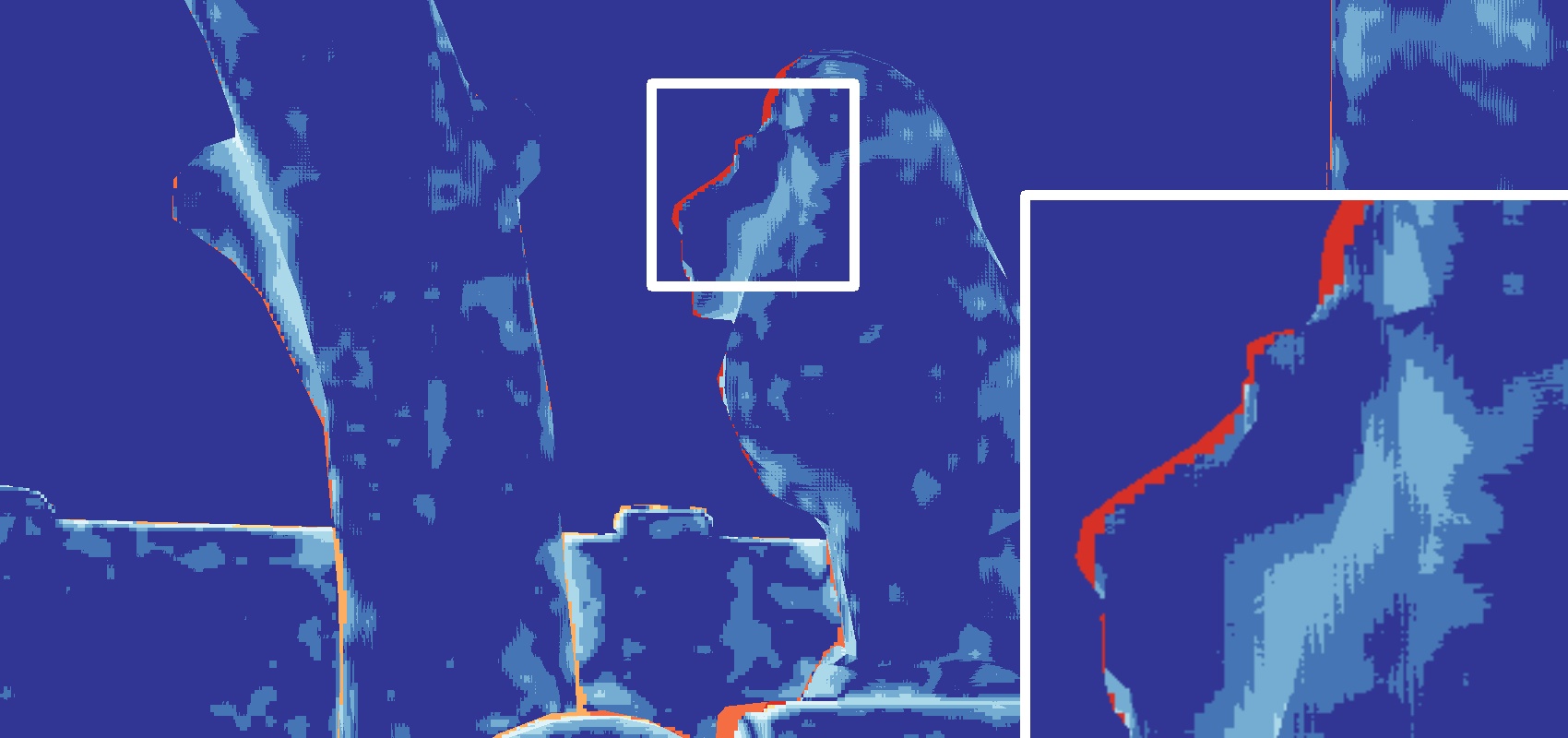} &
        \includegraphics[width=0.24\textwidth]{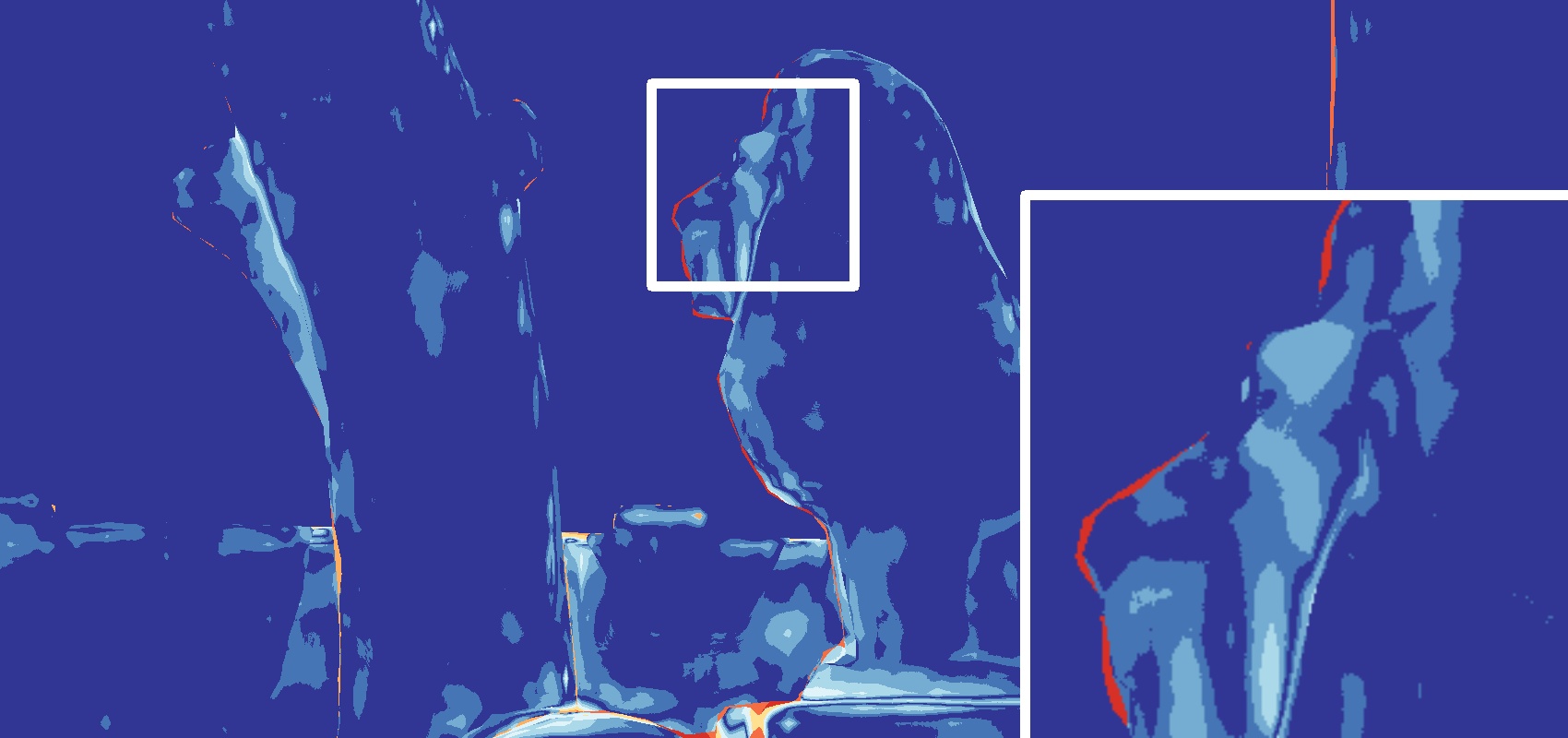}  &
        \includegraphics[width=0.24\textwidth]{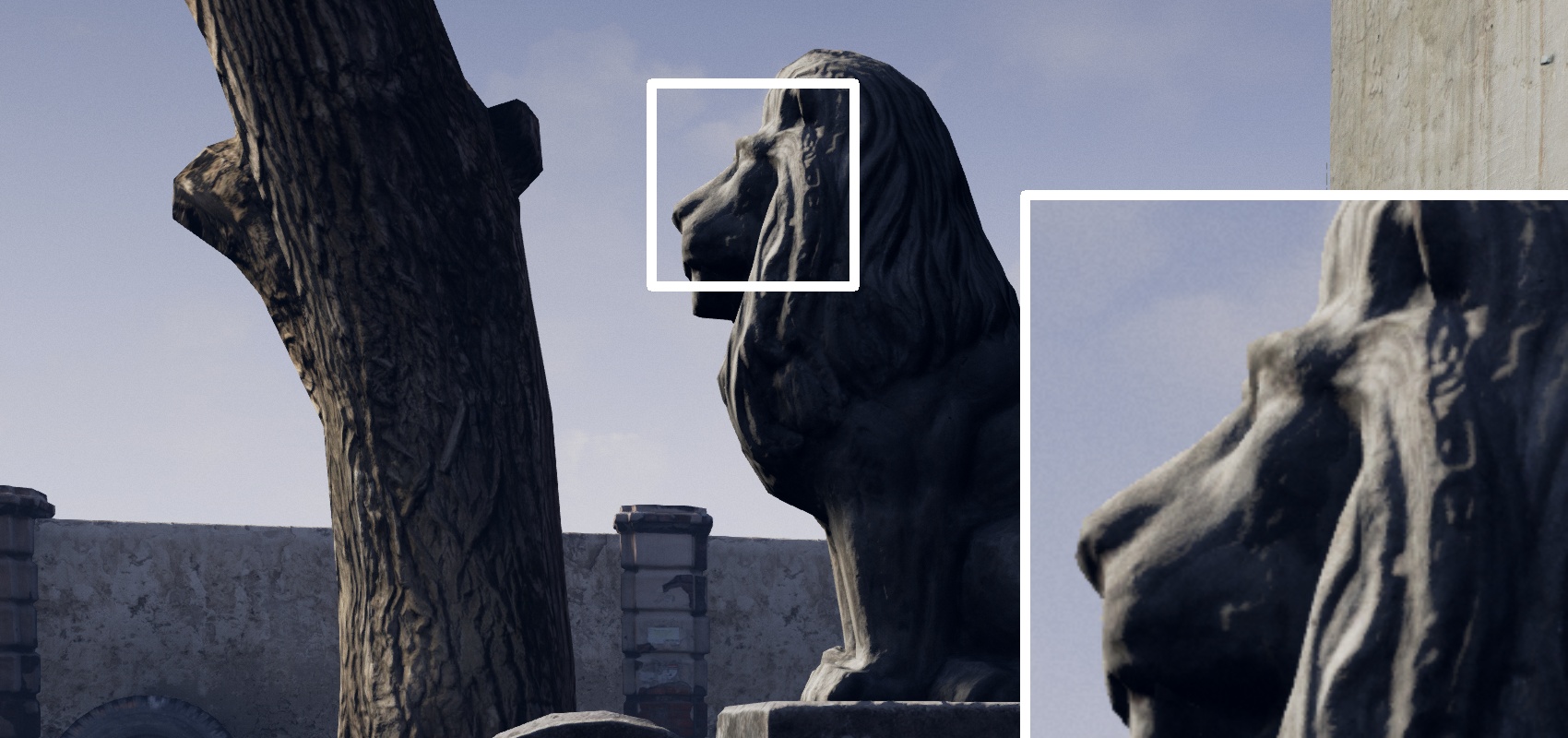} 
        \\
        \small{(a) PSM \cite{chang2018pyramid}} & \small{(b) PSM \cite{chang2018pyramid} + CE + SM  \cite{chen2019over}} & (c) \small{PSM + Ours} & \small{(d) GT, Input}
    \end{tabular}
    \vspace{-0.3cm}
    \caption{\textbf{Qualitative Results on UnrealStereo4K.} The first row shows the predicted disparity maps while the second row depicts the corresponding error maps. We zoom-in a patch in all images to better perceive  details near depth boundaries. 
    }
    \vspace{-0.2cm}
    \label{fig:qualitative}
\end{figure*}

\boldparagraph{Sampling Strategy}  In \tabref{table:dilation_analysis}, we show the impact of the sampling strategy adopted during training. More specifically, we compare the na\"ive uniform sampling strategy and the proposed \textit{DDA} approach using different dilation kernel sizes $\rho \times \rho$. As can be observed, \textit{DDA} enables SMD-Nets to focus on depth discontinuities, resulting in better \SEE compared to random point selection. Moreover, we observe that sampling exactly at depth boundaries (\ie, $\rho = 0$) leads to slightly degraded \EPE and is less effective on \SEE which penalizes small misalignment in a local window. Instead, setting $\rho=10$ allows the network to focus on larger regions near edges and results in the best performance, while increasing $\rho$ does not improve performance further.
Finally, it is worth to notice that this strategy also allows our model to improve the overall performance, achieving lower \EPE metrics. In the following experiments, we thus adopt the \textit{DDA} strategy using $\rho=10$ for our SMD-Nets. 

\boldparagraph{Ground Truth Resolution} \tabref{table:superres} shows the results of our model trained and tested on the stereo data using ground truth maps at different resolutions, while maintaining the input size at $960 \times 540$. Towards this goal, we train our model adopting ground truth disparities 1) resized to the same resolution as the input using nearest interpolation and 2) at the original resolution (\ie $3840 \times 2160$). 
We notice that sampling points from higher resolution disparity maps always leads to  better results compared to using low resolution ground truth. We remark that the proposed model effectively leverages high resolution ground truth thanks to its continuous formulation, without requiring additional memory compared to standard stereo networks based on CNNs.

\subsection{Comparison to Existing Baselines}

We now compare to several baselines \cite{Ma2013ICCV,chen2019over} which aim to address the over-smoothing problem.
Bilateral median filtering (BF) is often adopted to sharpen disparity predictions~\cite{Ma2013ICCV,Shih3DP20}.
Chen et al.~\cite{chen2019over} address the over-smoothing problem of 3D stereo backbones using 1) a post-processing step to extract a single-modal (SM) distribution from the full discrete distribution; 2) a cross-entropy (CE) loss to enforce a unimodal distribution during training. We reimplement \cite{chen2019over} as no official code is available.
As \cite{chen2019over} has been proposed for 3D backbones only, we use PSM \cite{chang2018pyramid} and HSM \cite{yang2019hierarchical} as the stereo backbones in the following experiments.

\boldparagraph{UnrealStereo4K} \tabref{table:unreal_comparison} collects results obtained from different models on both \textit{in-domain} and \textit{out-of-domain} test splits of the binocular UnrealStereo4K dataset.
We use the same input resolution of $960\times 540$ for all methods. While our baseline methods can only use supervision with the same size as the input, we leverage our continuous formulation to supervise SMD-Nets using ground truth at $3840\times 2160$, on which we also evaluate all methods.
For our competitors, we upsample their outcome using nearest neighbor interpolation during testing.  Both original PSM and HSM follow the same training setting of our SMD-Nets.

\tabref{table:unreal_comparison} suggests that BF~\cite{Ma2013ICCV} and SM~\cite{chen2019over} slightly improve  \SEE on both backbones while leading to degraded performance on \EPE metrics. Using the CE loss combined with SM~\cite{chen2019over} leads to effective improvement on both \SEE and \EPE on the PSM backbone. Interestingly, we notice that adopting the same CE + SM strategy leads to worse performance on HSM.
A possible explanation is that the CE loss requires to trilinearly interpolate a matching cost probability distribution to the full resolution $W\times H \times D_{max}$ (with $D_{max}$ denoting the maximum disparity), where HSM predicts a less fine-grained cost distribution compared to PSM, thus making the cross-entropy loss less effective.
Moreover, we remark that the CE loss is more expensive to compute, compared to our simple continuous likelihood-based formulation and CE + SM can only be applied on 3D backbones. In contrast, our approach based on the bimodal output representation notably outperforms our competitors on \SEE on both the \textit{in-domain} and \textit{out-of-domain} test sets, showing how our strategy predicts better disparities near boundaries. Moreover, we highlight that we achieve consistently better estimates on standard \EPE metrics compared to the original backbone while performing comparably to the CE + SM baseline. \figref{fig:qualitative} shows our gains at object boundaries.

\input{tables/realworld_ablation}
\input{tables/generalization}

\boldparagraph{KITTI 2015} We fine-tune all methods trained using UnrealStereo4K on the KITTI 2015 training set. Since the provided ground truth disparities are sparse, we rely on the na\"ive random sampling strategy to train our model. On the validation set, we evaluate \SEE on boundaries of instance segmentation maps from the KITTI dataset, following the evaluation procedure described in \cite{chen2019over}.  Furthermore, we predict disparities on the test set using the same fine-tuned model and submit to the online benchmark.  \tabref{table:kitti_internal} and \tabref{table:kitti_submission} show our results using PSM as backbone (we provide additional results on the validation set adopting HSM in the supplement). 
Note that our SMD-Net not only achieves superior performance on both \SEE and \EPE metrics on the validation set compared to the original PSM and \cite{chen2019over} (\tabref{table:kitti_internal}), but also outperforms both on the test set and is on par with state-of-the-art stereo networks on standard metrics of the KITTI benchmark (\tabref{table:kitti_submission}).

\subsection{Synthetic-to-Real Generalization}

Lastly, we demonstrate how models trained on the synthetic dataset generalize to the real-world domain for both binocular stereo and active depth estimation. 

\boldparagraph{Middlebury v3} \tabref{table:middlebury_exp} reports the performance of supervised methods trained on the UnrealStereo4K and tested without fine-tuning on the training set of the Middlebury v3 dataset. We evaluate them using the high-resolution ground truth. Compared to the original PSM baseline, our SMD-Net achieves much better generalization on both \textit{SEE} and \textit{EPE} metrics while performing on par with \cite{chen2019over}.

\begin{figure}
    \centering
    \begin{subfigure}{0.45\linewidth}
        \centering
        \includegraphics[width=\linewidth]{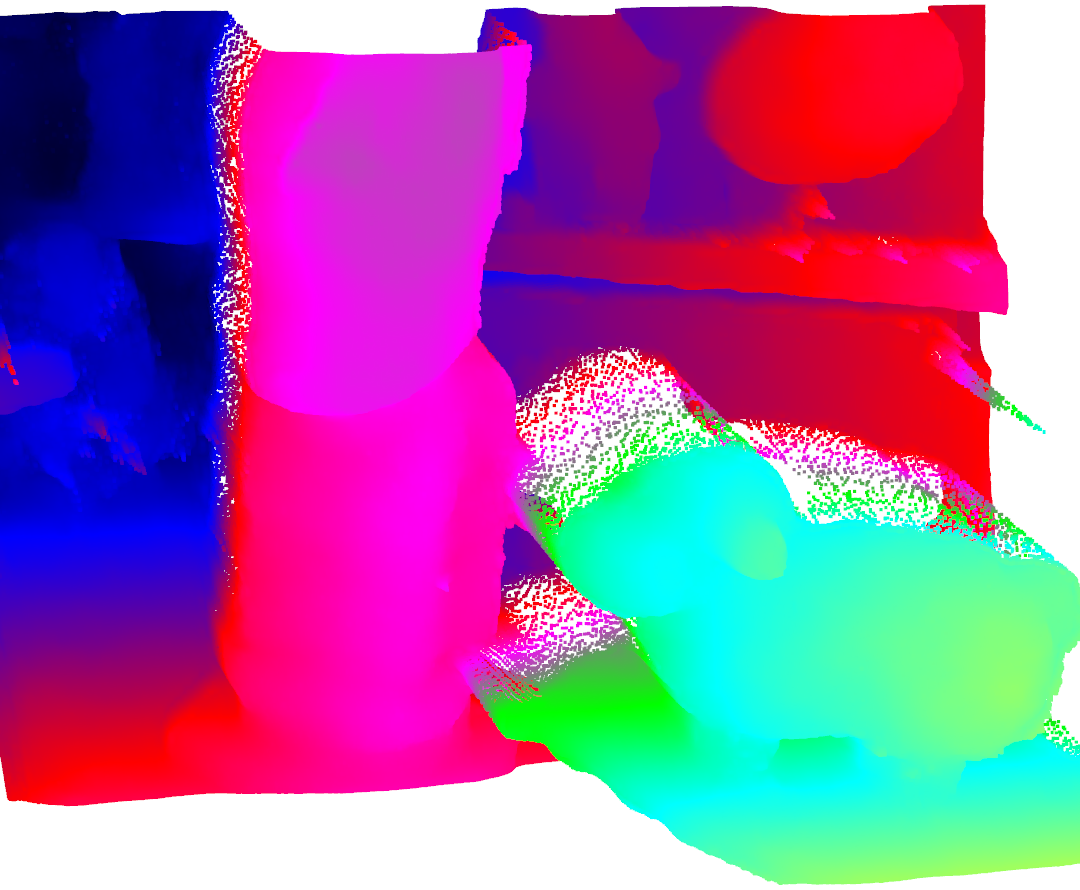}
        \caption{\small{Disparity Regression (L1)}}
        \label{fig:active_l1}
    \end{subfigure}
    \hfill
    \begin{subfigure}{0.45\linewidth}
        \centering
        \includegraphics[width=\linewidth]{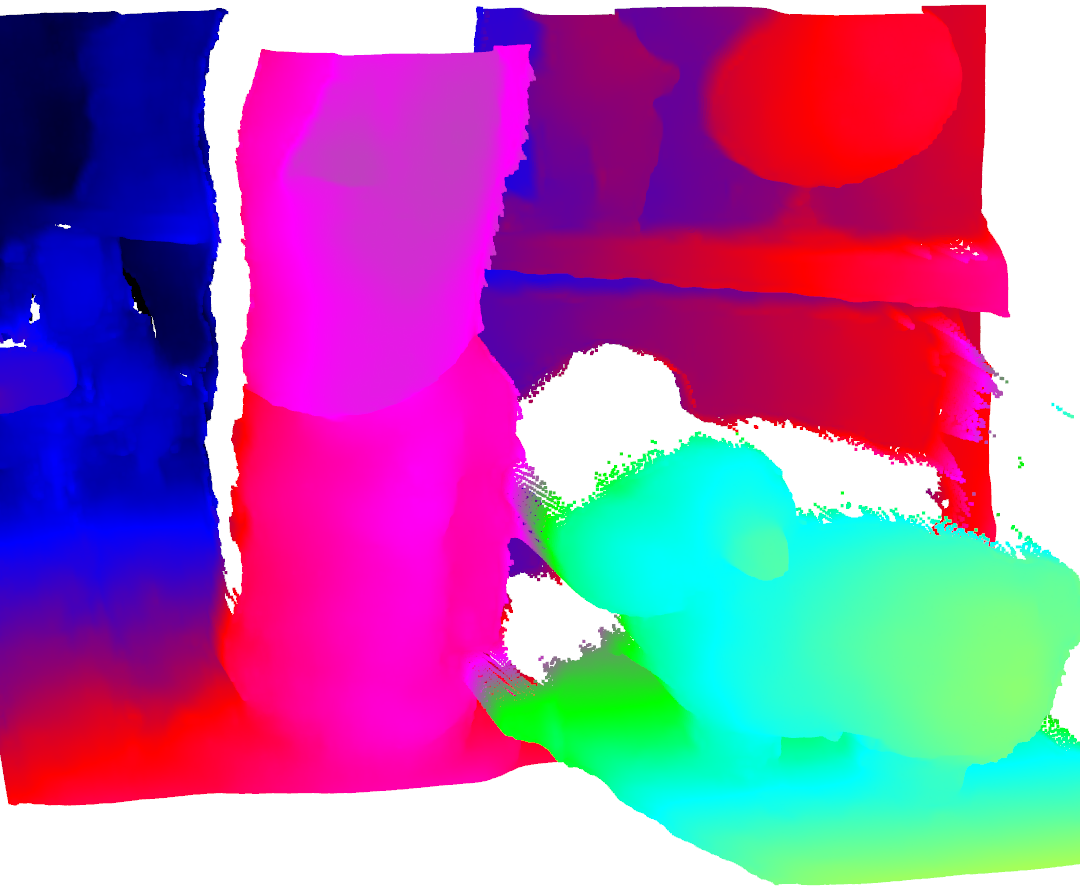}
        \caption{\small{SMD Head (Bimodal)}}
        \label{fig:active_bimodal}
    \end{subfigure}
    \vspace{-0.2cm}
    \caption{\textbf{Generalization on RealActive4K} using the HSM backbone. The point clouds of standard disparity regression using L1 loss (\subref{fig:active_l1}) show bleeding artifacts whereas our bimodal distribution (\subref{fig:active_bimodal}) leads to clean reconstructions.} 
    \label{fig:active_real}
    \vspace{-0.4cm}
\end{figure}

\boldparagraph{RealActive4K} Moreover, we fine-tune our active depth models jointly on active UnrealStereo4K and RealActive4K with pseudo-ground truth from Block Matching. \figref{fig:active_real} shows that this allows for estimating sharp disparity edges for real captures even though Block Matching does not provide supervision in these areas. In contrast, standard disparity regression fails to predict clean object boundaries.

%% file: tables/unreal_ablation2.tex
\begin{table}[tp]
\centering
\scalebox{0.67}{
\begin{tabular}{l|c|c|ccc|ccc|cc}
\hline
& \multirow{2}{*}{$\Psi_{\theta}$} & \multirow{2}{*}{Dim.}  & \multicolumn{3}{c|}{$\SEEk{3}$} & \multicolumn{3}{c|}{$\SEEk{5}$} & \multicolumn{2}{c}{\EPE} \\
\cline{4-11}
& & & Avg & $\sigma(1)$ & $\sigma(2)$ & Avg & $\sigma(1)$ & $\sigma(2)$ & Avg & $\sigma(3)$\\
\hline

\multirow{6}{*}{\rotatebox[origin=c]{90}{\textit{Binocular Stereo}}} &
\multirow{3}{*}{\begin{minipage}{0.9cm}\centering U-Net \\ (2D ) \\ \cite{monodepth2} \end{minipage}}  & 1 & 2.15 & 41.69 & 24.16 & 2.03 & 39.65 & 22.98 & 1.48 & 8.18\\
& & 2  &  2.38 & 42.28 & 25.74 & 2.26 & 40.42 & 24.57 & 1.97 & 10.44\\
& & 5 & \textbf{1.57} & \textbf{30.06} & \textbf{14.77} & \textbf{1.45} & \textbf{28.05} & \textbf{16.57} & \textbf{1.28} & \textbf{5.94}\\
\cline{2-11}

& \multirow{3}{*}{\begin{minipage}{0.9cm}\centering PSM \\ (3D) \\  \cite{chang2018pyramid} \end{minipage}} & 1 & 1.98 & 36.32 & 20.35 & 1.85 & 34.42 & 19.21 & \textbf{1.10} & 5.52\\
& & 2 & 2.50 & 39.40 & 23.63 & 2.37 & 37.57 & 22.51 & 1.88 & 7.73 \\
& & 5 & \textbf{1.52} & \textbf{26.98} & \textbf{12.68} & \textbf{1.38} & \textbf{24.93} & \textbf{11.49}  & \textbf{1.11} & \textbf{4.80}\\
\hline

\hline
\multirow{3}{*}{\rotatebox[origin=c]{90}{\textit{Mono.}}} &
\multirow{3}{*}{\begin{minipage}{0.9cm}\centering U-Net \\ (2D) \cite{monodepth2} \end{minipage}} & 1  & 3.29 & 60.18 & 41.37 & 3.25 & 58.49 & 40.08 & 4.21 & 35.92 \\
& & 2  & 4.01 & 61.06 & 43.19 & 3.86 & 59.40 & 41.90 & 5.49 & 41.88\\
& & 5 & \textbf{2.92} & \textbf{51.32} & \textbf{32.33} & \textbf{2.78} & \textbf{49.54} & \textbf{31.06} & \textbf{4.06} & \textbf{30.59}\\
\hline

\hline
\multirow{3}{*}{\rotatebox[origin=c]{90}{\textit{Active}}}  &
\multirow{3}{*}{\begin{minipage}{0.9cm}\centering HSM \\ (3D) \cite{yang2019hierarchical} \end{minipage}} & 1 & 3.40 & 47.87 & 24.80 & 3.18 & 46.14 & 23.76 & \textbf{1.29} & 5.84  \\
& & 2  & 4.93 & 57.05 & 33.44 & 4.69 & 55.47 & 32.41 & 2.83 & 10.70\\
& & 5 & \textbf{2.69} & \textbf{41.84} & \textbf{17.35} & \textbf{2.43} & \textbf{39.83} & \textbf{16.17} & 1.42 & \textbf{5.48}\\
\hline

\end{tabular}
} 
\vspace{-0.2cm}
\caption{\textbf{Output Representation} analysis on the UnrealStereo4K test set. ``Dim.'' refers to the output dimension of the SMD Head where 1 indicates the point estimate $d$, 2  the unimodal output representation $(\mu, b)$ \cite{kendall2017uncertainties} and 5 our bimodal formulation $(\pi,\mu_{1}, b_{1},\mu_{2}, b_{2})$.}
\label{table:ablation}
\end{table}

%% file: tables/dilation_factor.tex
\begin{table}[tp]
\centering
\scalebox{0.67}{
\setlength\tabcolsep{0.68em}
\begin{tabular}{cc|ccc|ccc|cc}
\hline
\multirow{2}{*}{Sampling} & \multirow{2}{*}{$\rho$}  & \multicolumn{3}{c|}{$\SEEk{3}$} & \multicolumn{3}{c|}{$\SEEk{5}$} & \multicolumn{2}{c}{\EPE} \\
\cline{3-10}
& & Avg & $\sigma(1)$ & $\sigma(2)$ & Avg & $\sigma(1)$  & $\sigma(2)$  & Avg & $\sigma(3)$\\
\hline
Random & - & 1.52 &  26.98 & 12.68 & 1.38 & 24.93 & 11.49 & 1.11 & 4.80 \\
DDA & 0 &  1.34 & 21.62 & 9.77 & 1.19 & 19.58 & 8.59 & 1.08 & 4.44\\
DDA & 10 & \textbf{1.13} & \textbf{18.64} & \textbf{8.69} & \textbf{0.98} & \textbf{16.67} & \textbf{7.55} & \textbf{0.92} & \textbf{3.88}\\
DDA & 20 & 1.30 & 20.42 & 9.88 & 1.15 & 18.40 & 8.71 & 1.11 & 4.44\\
\hline
\end{tabular}
}
\vspace{-0.2cm}
\caption{\textbf{Sampling Strategy} analysis on the UnrealStereo4K test set using the PSM backbone.}
\label{table:dilation_analysis}
\end{table}

%% file: tables/superresolution.tex
\begin{table}[tp]
\centering
\scalebox{0.67}{
\setlength\tabcolsep{0.45em}
\begin{tabular}{cc|ccc|ccc|cc}
\hline
\multirow{2}{*}{Eval. GT}  & \multirow{2}{*}{Training GT} & \multicolumn{3}{c|}{$\SEEk{3}$} & \multicolumn{3}{c|}{$\SEEk{5}$} & \multicolumn{2}{c}{\EPE} \\
\cline{3-10}
& & Avg & $\sigma(1)$  & $\sigma(2)$  & Avg & $\sigma(1)$  & $\sigma(2)$  & Avg & $\sigma(3)$\\
\hline
\small{$960 \times 540$}  & \small{$960 \times 540$}   & 1.19 & 20.36 & 9.16 & 0.93 & 16.55 & 7.08 & 1.02 & 4.30\\
\small{$960 \times 540$}  &  \small{$3840\times 2160$} & \textbf{0.98 }& \textbf{15.42} & \textbf{7.05} & \textbf{0.78} & \textbf{12.44} & \textbf{5.54} & \textbf{0.89} & \textbf{3.81}\\
\hline
\small{$3840\times 2160$}  & \small{$960 \times 540$}   & 1.33 & 23.35 & 10.82 & 1.19 & 21.34 & 9.63 & 1.03 & 4.30\\
\small{$3840\times 2160$} & \small{$3840 \times 2160$} &  \textbf{1.13} & \textbf{18.64} & \textbf{8.69} & \textbf{0.98} & \textbf{16.67} & \textbf{7.55} & \textbf{0.92} & \textbf{3.88}\\
\hline
\end{tabular}
}
\vspace{-0.2cm}
\caption{\textbf{Ground Truth Resolution} analysis on the UnrealStereo4K test set using the PSM backbone.}
\label{table:superres}
\vspace{-0.3cm}
\end{table}

%% file: tables/unreal_comparison.tex
\begin{table*}[!htbp]
\centering
\scalebox{0.65}{
\setlength\tabcolsep{0.57em}
\begin{tabular}{l|ccc|ccc|cccc||ccc|ccc|cccc}
\hline
 \multirow{3}{*}{Method} & \multicolumn{10}{c||}{\textit{In-domain}} &
 \multicolumn{10}{c}{\textit{Out-of-domain}}
 \\
 \cline{2-21}
 & \multicolumn{3}{c|}{$\SEEk{3}$} & \multicolumn{3}{c|}{$\SEEk{5}$} & \multicolumn{4}{c||}{\EPE} & \multicolumn{3}{c|}{$\SEEk{3}$} & \multicolumn{3}{c|}{$\SEEk{5}$} & \multicolumn{4}{c}{\EPE} \\
\cline{2-21}
 & Avg & $\sigma(1)$  & $\sigma(2)$  & Avg & $\sigma(1)$  & $\sigma(2)$  & Avg & $\sigma(1)$  & $\sigma(2)$  & $\sigma(3)$  & Avg & $\sigma(1)$  & $\sigma(3)$  & Avg & $\sigma(1)$  & $\sigma(3)$  & Avg & $\sigma(1)$  & $\sigma(2)$  & $\sigma(3)$  \\
\hline
PSM \cite{chang2018pyramid} 
& 1.73 & 33.06 & 16.57 & 1.61 & 31.11 & 15.44 & 1.09 & 11.88 & 6.94 & 5.19  
& 2.19 & 36.94 & 20.07 & 1.99 & 34.09 & 18.25 & 1.53 & 16.92 & 10.25 & 7.83 
\\
PSM \cite{chang2018pyramid} + BF \cite{Ma2013ICCV} 
& 1.65 & 30.93 & 15.26 & 1.52 & 28.92 & 14.10 & 1.10 & 11.81 & 6.95 & 5.23 
& 2.16 & 35.64 & 19.16 & 1.95 & 32.76 & 17.32 & 1.56 & 18.89 & 10.28 & 7.89 
\\
PSM \cite{chang2018pyramid} + SM \cite{chen2019over} 
& 1.50 & 29.22 & 12.71 & 1.37 & 27.16 & 11.54 & 1.10 & 11.65 & 6.69 & 4.97  
& 2.03 & 33.91 & 16.74 & 1.82 & 30.92 & 14.82 & 1.54 & 16.43 & 9.73 & 7.36  
\\
PSM \cite{chang2018pyramid} + CE + SM \cite{chen2019over} 
& 1.33 & 27.31 & 10.14 & 1.19 & 25.25 & 8.99 & \textbf{0.86} & 10.40 & \textbf{4.93} & \textbf{3.50} 
& 1.84 & 29.87 & 13.30 & 1.62 & 26.84 & 11.46 & 1.37 & 13.29 & 7.84 & \textbf{6.03} 
\\
PSM \cite{chang2018pyramid} + Ours 
& \textbf{1.13} & \textbf{18.64} & \textbf{8.69} & \textbf{0.98} & \textbf{16.67} & \textbf{7.55} & 0.92 & \textbf{8.24} & 5.06 & 3.88  
& \textbf{1.59} & \textbf{24.58} & \textbf{12.54} & \textbf{1.38} & \textbf{21.63} & \textbf{10.73} & \textbf{1.27} & \textbf{12.11} & \textbf{7.69} & \textbf{6.06}  
\\
\hline
HSM \cite{yang2019hierarchical}   
& 2.01 & 41.63 & 23.81 & 1.89 & 39.69 & 22.62 & 1.16 & 14.81 & 8.20 & 5.84  
& 2.43 & 44.49 & 26.17 & 2.24 & 41.74 & 24.33 & 1.75 & 22.03 & 12.73 & 9.23 
\\
HSM \cite{yang2019hierarchical} + BF \cite{Ma2013ICCV} 
& 1.88 & 39.68 & 21.70 & 1.77 & 37.67 & 20.49 & 1.19 & 14.78 & 8.21 & 5.88 
& 2.39 & 43.60 & 24.14 & 2.19 & 40.82 & 23.28 & 1.80 & 22.05 & 12.79 & 9.33 
\\
HSM \cite{yang2019hierarchical} + SM \cite{chen2019over}  
& 1.83 & 40.52 & 22.30 & 1.70 & 38.53 & 21.07 & 1.17 & 14.73 & 8.11 & 5.74  
& 2.31 & 43.76 & 25.16 & 2.11 & 40.97 & 23.29 & 1.76 & 21.88 & 12.54 & 9.03 
\\
HSM \cite{yang2019hierarchical} + CE + SM \cite{chen2019over} 
& 2.00 & 45.71 & 25.99 & 1.87 & 43.72 & 24.71 & 1.17 & 16.17 & 8.12 &  5.46 
& 2.61 & 48.27  & 28.84 & 2.41 & 45.56 & 26.98 & 1.91 & 26.12 & 14.40 & 10.14 
\\
HSM \cite{yang2019hierarchical} + Ours
& \textbf{1.31} & \textbf{24.31} & \textbf{10.81} & \textbf{1.17} & \textbf{22.30} & \textbf{9.67} & \textbf{1.00} & \textbf{11.40} & \textbf{6.09} & \textbf{4.34} 
& \textbf{2.03} & \textbf{34.82} & \textbf{17.75} & \textbf{1.82} & \textbf{31.88} & \textbf{15.83} & \textbf{1.66} & \textbf{19.16} & \textbf{10.72} & \textbf{7.77}
\\
\hline
\end{tabular}
}
\vspace{-0.2cm}
\caption{\textbf{Comparison on UnrealStereo4K.} All methods evaluated on ground truth at  $3840 \times 2160 $ given input size $960 \times 540$.}
\label{table:unreal_comparison}
\end{table*}

%% file: tables/realworld_ablation.tex
\begin{table}[tp]
\centering
\scalebox{0.63}{
\begin{tabular}{l|ccc|ccc|cc}
\hline
 \multirow{2}{*}{Method}  &\multicolumn{3}{c|}{$\SEEk{3}$} & \multicolumn{3}{c|}{$\SEEk{5}$} & \multicolumn{2}{c}{\EPE} \\
\cline{2-9}
& Avg & $\sigma(1)$  & $\sigma(2)$  & Avg & $\sigma(1)$ & $\sigma(2)$  & Avg & $\sigma(3)$ \\
\hline
PSM \cite{chang2018pyramid} & 1.10 & 20.57 & 9.74 & 0.99 & 17.83 & 9.02 & 0.73 & 2.49\\
PSM \cite{chang2018pyramid} + CE + SM \cite{chen2019over} &1.02 & 16.12 & 7.53 & 0.90 & 13.80 & 6.94 & 0.66 & 2.09 \\
PSM \cite{chang2018pyramid} + Ours & \textbf{0.90} & \textbf{13.09} & \textbf{6.66} & \textbf{0.79} & \textbf{10.93} & \textbf{6.01} & \textbf{0.59} & \textbf{1.95}\\
\hline
\end{tabular}
}
\vspace{-0.2cm}
\caption{\textbf{Comparison on KITTI 2015 Validation Set} using boundaries extracted from instance segmentation masks to evaluate on depth discontinuity regions.}
\label{table:kitti_internal}
\end{table}
\begin{table}[tp]
\centering
\setlength\tabcolsep{1.1em}
\scalebox{0.65}{
\begin{tabular}{l|ccc|ccc}
\hline
 \multirow{2}{*}{Method}  &\multicolumn{3}{c|}{All Areas} & \multicolumn{3}{c}{Non Occluded} \\
\cline{2-7}
& Bg & Fg  & All  & Bg & Fg & All\\
\hline
GANet-deep \cite{zhang2019ga} & 1.48 & 3.46 & 1.81 & 1.34 & 3.11 & 1.63\\
HD$^3$-Stereo \cite{yin2019hierarchical} & 1.70 & 3.63 & 2.02 & 1.56 & 3.43 & 1.87\\
GwcNet-g \cite{guo2019group} & 1.74  & 3.93 & 2.11 & 1.61 & 3.49 & 1.92\\
\hline
PSM \cite{chang2018pyramid} & 1.86 & 4.62 & 2.31 & 1.71 & 4.31 & 2.14\\
PSM \cite{chang2018pyramid} + CE + SM \cite{chen2019over}  & \textbf{1.54} & 4.33 & 2.14 & 1.70 & 3.90 & 1.93\\
PSM \cite{chang2018pyramid} + Ours & 1.69 & \textbf{4.01} & \textbf{2.08} & \textbf{1.54} & \textbf{3.70} & \textbf{1.89}\\
\hline
\end{tabular}
}
\vspace{-0.2cm}
\caption{\textbf{Comparison on KITTI 2015 Test Set}, evaluated on the official online benchmark. All the reported numbers represent official submissions from the authors. }
\label{table:kitti_submission}
\end{table}

%% file: tables/generalization.tex
\begin{table}[tp]
\centering
\scalebox{0.63}{
\setlength\tabcolsep{0.53em}
\begin{tabular}{l|ccc|ccc|cc}
\hline
 \multirow{2}{*}{Method} &\multicolumn{3}{c|}{$\SEEk{3}$} & \multicolumn{3}{c|}{$\SEEk{5}$} & \multicolumn{2}{c}{\EPE} \\
\cline{2-9}
& Avg & $\sigma(1)$  & $\sigma(2)$  & Avg & $\sigma(1)$  & $\sigma(2)$  & Avg & $\sigma(3)$ \\
\hline
PSM \cite{chang2018pyramid} & 3.35 & 46.50 & 29.40 & 2.61 & 41.04 & 24.87 & 4.12 & 17.43\\
PSM \cite{chang2018pyramid} + CE + SM \cite{chen2019over} & \textbf{2.62} & 34.80 & \textbf{19.02} & \textbf{1.83} & 28.92 & \textbf{14.11} & \textbf{2.80} & \textbf{12.12} \\
PSM \cite{chang2018pyramid} + Ours  & \textbf{2.61} & \textbf{34.26} & 19.83 & \textbf{1.88} & \textbf{28.71} & 15.32 & 3.03 & 13.60\\
\hline
\end{tabular}
}
\vspace{-0.2cm}
\caption{\textbf{Generalization on Middlebury v3}. All models are trained on UnrealStereo4K and evaluated on the training set of Middlebury v3 dataset.}
\label{table:middlebury_exp}
\vspace{-0.3cm}
\end{table}

%% file: 5_conclusion.tex
\section{Conclusion}

In this paper, we propose SMD-Nets, a novel stereo matching framework aimed at improving depth accuracy near object boundaries and suited for disparity super-resolution. By exploiting bimodal mixture densities as output representation combined with a continuous function formulation, our method is capable of predicting sharp and precise disparity values at arbitrary spatial resolution, notably alleviating the common over-smoothing problem in learning-based stereo networks. Our model is compatible with a broad spectrum of 2D and 3D stereo backbones. Our extensive experiments demonstrate the advantages of our strategy on a new high-resolution synthetic stereo dataset and on real-world stereo pairs. We plan to extend our bimodal output representation to other regression tasks such as optical flow and self-supervised depth estimation.\\